\documentclass[conference]{IEEEtran}
\IEEEoverridecommandlockouts
\usepackage{cite}
\usepackage{amsmath,amssymb,amsfonts}
\usepackage{algorithmic}
\usepackage{adjustbox}
\usepackage{graphicx}
\graphicspath{ {./images/} }
\usepackage{amssymb}% http://ctan.org/pkg/amssymb
\usepackage{pifont}% http://ctan.org/pkg/pifont
\newcommand{\cmark}{\ding{51}}

\usepackage{textcomp}
\usepackage{xcolor}
\usepackage{multirow,tabularx}
\usepackage{alphalph}
\usepackage{caption}
\usepackage{comment}

\renewcommand\thesubsectiondis{\AlphAlph{\value{subsection}}.}
\renewcommand\thesubsection{\mbox{\thesection-\AlphAlph{\value{subsection}}}}

\def\BibTeX{{\rm B\kern-.05em{\sc i\kern-.025em b}\kern-.08em
    T\kern-.1667em\lower.7ex\hbox{E}\kern-.125emX}}
\begin{document}

\title{Camera-Radar Perception for Autonomous Vehicles and ADAS: Concepts, Datasets and Metrics}

\author{\IEEEauthorblockN{1\textsuperscript{st}Felipe Manfio Barbosa}
\IEEEauthorblockA{\textit{Institute of Mathematical and Computer Sciences (ICMC)} \\ University of S$\tilde{a}$o Paulo\\
S$\tilde{a}$o Carlos, S$\tilde{a}$o Paulo
%\\
%Email: felipe.manfio.barbosa@usp.br
}
\and
\IEEEauthorblockN{2\textsuperscript{nd} Fernando Santos Os$\acute{o}$rio}
\IEEEauthorblockA{\textit{Institute of Mathematical and Computer Sciences (ICMC)} \\ University of S$\tilde{a}$o Paulo\\
S$\tilde{a}$o Carlos, S$\tilde{a}$o Paulo
%\\
%Email: fosorio@icmc.usp.br
}
}

\begin{comment}
\author{
  {1\textsuperscript{st} Felipe Manfio Barbosa}\qquad
  {2\textsuperscript{nd} Fernando Santos Os$\acute{o}$rio}\qquad
  {3\textsuperscript{rd} Rodolfo Ipolito Meneguette}  \\
  $^1${University of São Paulo}\\
  {\tt\small felipe.manfio.barbosa@usp.com\qquad
            \{fosorio,meneguette\}@icmc.usp.com}
}
\end{comment}

% Institute of Mathematics and Computer Science 
% Department of Computer Systems

\maketitle

\begin{abstract}
One of the main paths towards the reduction of traffic accidents is the increase in vehicle safety through driver assistance systems or even systems with a complete level of autonomy. In these types of systems, tasks such as obstacle detection and segmentation, especially the Deep Learning-based ones, play a fundamental role in scene understanding for correct and safe navigation. Besides that, the wide variety of sensors in vehicles nowadays provides a rich set of alternatives for improvement in the robustness of perception in challenging situations, such as navigation under lighting and weather adverse conditions. Despite the current focus given to the subject, the literature lacks studies on radar-based and radar-camera fusion-based perception. Hence, this work aims to carry out a study on the current scenario of camera and radar-based perception for ADAS and autonomous vehicles. Concepts and characteristics related to both sensors, as well as to their fusion, are presented. Additionally, we give an overview of the Deep Learning-based detection and segmentation tasks, and the main datasets, metrics, challenges, and open questions in vehicle perception.

\end{abstract}


\section{Introduction}
Traffic safety is an issue of fundamental importance in everyday life around the world. According to data released by the World Health Organization in the Global Status Report on Road Safety (2018) \cite{b1}, traffic accidents result, annually, in the death of approximately 1.35 million people worldwide, being the 8th most common cause of death among people of all ages and the leading cause of death among children and young adults, aged between 5 and 29 years. Additionally, more than half (54\%) of all traffic-related deaths and injuries involve vulnerable road users, such as pedestrians, cyclists, motorcyclists, and their passengers.
\par From a socio-economic point of view, traffic accidents represent expenses of almost 3\% of the Gross Domestic Product (GDP) of most countries analyzed. 
\par It has an even worse impact in developing countries. Despite representing 60\% of the world's vehicle fleet, low and middle-income countries account for 93\% of the traffic accidents, with fatal accident rates three times higher than high-income countries.
\par In addition to all these factors, there is the immeasurable damage to the families of traffic victims.
\par These alarming statistics motivated public and private efforts towards the development of action plans for road safety. An example of such an initiative is the Second Decade of Action for Traffic Safety \cite{b2} that, between the years 2021 to 2030, aims to reduce at least 50\% of road traffic injuries and deaths worldwide.
\par Despite its current notability, vehicle safety has been a subject of research for a long time. Over the years, the scientific community and the market have made relevant advances in the area, such as the development of Driver Assistance Systems (DAS), Advanced Driver Assistance Systems (ADAS), and, more recently, Autonomous Vehicles. 
\par ADAS can have active or passive actuation. In the first case, the system performs punctual decisions and actions to avoid risky situations based on its perception of the environment. In passive operation, on the other hand, the system just informs about possible hazards to the driver who is, thus, responsible for acting to prevent accidents.
\par Autonomous vehicles, in turn, are capable of navigating the environment the entire time without the intervention of a human agent. This level of autonomy is achieved through an integration of the vehicle's perception, decision-making and actuation systems. They can also be classified as critical systems, since errors resulting from failures in some of their modules can pose serious material and life risks to people in the environment. 
\par As an example, in 2018 an Uber autonomous car operating in low light conditions killed a cyclist due to an error in its vision system. Because the vehicle was not able to correctly recognize the cyclist, it incorrectly calculated its trajectory and the time to activate the brakes \cite{uber_accident}.
\par This demonstrates how perception can be a complex task in adverse navigation conditions, commonly related to factors such as lighting - night navigation or sun glare - or weather - fog, dust, rain, snow. Therefore, it is crucial to consider such issues in studies related to autonomous vehicles and driver assistance systems.
% \par It also points to the importance of simulation to the development of vehicle perception. It allows data gathering, implementation, and testing without any inherent risk or the need for expensive sensing setups.
\par Camera-based perception is strongly influenced by such adverse conditions. Thus, although image data have been the basis for the great part of the advances in Computer Vision in recent years, it is necessary to study new perception strategies. In this context, the use of radar data and the study of hybrid sensor fusion have gained special attention as robust alternatives for perception in challenging conditions.
\par However, despite the advantages offered by such approaches, the literature still lacks studies on them.
\par The reduced availability of surveys covering the previously mentioned aspects, and the most recent relevant contributions in terms of camera, radar, and fusion-based perception, is a problem that must be addressed. 
This makes it difficult for beginners in vehicle perception to have a comprehensive starting point to the main concepts, datasets, metrics, and challenges of the field.

\subsection{Scope and Contributions}
Motivated by the richness of the field, and the fast pace at which new developments are proposed, the literature presents several works aimed at summarising the main concepts and contributions in perception for Autonomous Vehicles and ADAS. Those works are of primary importance for introducing new researchers to the field. \par In this survey, we approach the main concepts, datasets, and metrics used in camera and radar-based perception for autonomous vehicles and ADAS. 
\par First, we conceptualize ADAS and Autonomous Vehicles. Then, we analyze the pros and cons of each sensing modality used in vehicle perception and discuss the sensor fusion approach as a solution to their limitations. After that, the main concepts and some of the most prominent works in Deep Learning-based object detection and image segmentation are presented. Then, we summarize the main datasets and metrics used. Finally, we discuss the main challenges and make considerations about future directions for development in the field.
\par The main contributions of this work are as follows:

\begin{itemize}
  \item A comprehensive description of the camera and radar operation and characteristics, with a discussion of its main advantages and drawbacks in the context of autonomous vehicles and ADAS. We additionally present the sensor fusion approach as a way to overcome the limitations presented by each sensing modality individually;
  \item A comprehensive overview on the main concepts and Deep Learning-based methods for object detection and image segmentation for autonomous driving;
  \item A thorough aggregation of the most relevant and recent open-source datasets for vehicle perception. We present their main characteristics in terms of sensing modalities, data diversity, size, and intended perception task;
  \item A consolidation of the main metrics used to assess detection and segmentation performance - grouped by dataset;
  \item A discussion of the main challenges and future directions in perception for autonomous vehicles and ADAS, with a particular focus on the rule of radar perception in this context.
\end{itemize}

\subsection{Comparison with Existing Surveys}
The literature on vehicle perception, although very complete in terms of concepts, methods, and datasets, still lack in-depth exploration of some very important sensing modalities, such as radar-based perception. Additionally, to the best of our knowledge, the most recently proposed datasets are not covered by the surveys in the area.
\par Many works provide a complete consolidation of most of the available open-source datasets used in the context of autonomous vehicles and ADAS. The survey proposed in \cite{survey1} covers a wide range of publicly available driving datasets. However, the criteria used for work selection excluded datasets created via simulation. Additionally, no radar-based datasets have been reported. In \cite{survey2} it is performed a thorough analysis of multi-modal perception for object detection and semantic segmentation for autonomous driving. Although covering a wide range of methods, including the fusion-based ones, they are mainly camera and LIDAR-based, with the authors briefly describing radar sensing.
Additionally, none of them cover the newest datasets in the field - proposed after their publication -, neither the metrics used to assess Deep Learning-based object detection and image segmentation.
\par Other works focus their attention in describing the main methods used in perception tasks. The survey \cite{survey3} presents the datasets, metrics and methods used in general scene labeling, detailing the main architecture choices, frameworks and techniques used in Deep Learning-based scene labeling. In  the works \cite{survey4} and \cite{survey5}, the authors concentrate on general object detection. The main datasets and metrics used in 2D Deep Learning-based generic object detection are presented. Additionally, it is carried out an extensive historical review of Deep Learning-based generic object detection methods, as well as a discussion of future trends and challenges, with practical considerations on factors that can degrade detection performance.
Although very complete and useful studies on perception, these works focused only on camera data, and did not consider the context of autonomous vehicles and ADAS. Therefore, they do not study the use of radar data - or sensor fusion -, and also lack in making proper considerations about possible adverse operating conditions that camera-based detection can face.
\par Finally, the review \cite{survey7} heavily focuses on the radar and sensor fusion approaches for multi-object detection and tracking. Besides performing a rich discussion on driving conditions, the work provides important considerations on the different sensor's pros and cons, considering the context of autonomous vehicles' perception. However, it does not gathers neither the main datasets nor the metrics used in the field.

\begin{table*}
\caption{Contributions of this study compared to previous relevant works.}
\centering
\begingroup
\renewcommand{\arraystretch}{1.5} % Default value: 1
\begin{tabular}{ccccccccccccc}
\hline
\multirow{2}{*}{Survey} &
  \multirow{2}{*}{Year} &
  \multirow{2}{*}{\begin{tabular}[c]{@{}c@{}}AV\\ Context\end{tabular}} &
  \multirow{2}{*}{\begin{tabular}[c]{@{}c@{}}Sensing\\ Modalities\\ Description\end{tabular}} &
  \multicolumn{4}{c}{Sensing Modalities Covered} &
  \multirow{2}{*}{\begin{tabular}[c]{@{}c@{}}Data\\ Diversity\end{tabular}} &
  \multirow{2}{*}{Metrics} &
  \multirow{2}{*}{\begin{tabular}[c]{@{}c@{}}Num. of\\ Datasets\end{tabular}} &
  \multirow{2}{*}{\begin{tabular}[c]{@{}c@{}}Year\\ Range\end{tabular}} \\ \cline{5-8}
 &
   &
   &
   &
  \begin{tabular}[c]{@{}c@{}}Camera\\ (2D)\end{tabular} &
  \begin{tabular}[c]{@{}c@{}}Camera\\ (Stereo Vision)\end{tabular} &
  Radar &
  \begin{tabular}[c]{@{}c@{}}Sensor\\ Fusion\end{tabular} 
   \\ \hline
\cite{survey1} &
  2017 &
  \cmark &
   &
  \cmark &
  \cmark &
   &
   &
  \cmark &
   &
  27 &
  2006-2016 \\
\cite{survey2} &
  2020 &
  \cmark &
  \cmark &
  \cmark &
  \cmark &
  \cmark &
  \cmark &
  \cmark &
   &
  21 &
  2013-2019 \\
\cite{survey3} &
  2017 &
   &
   &
  \cmark &
  \cmark &
   &
   &
   &
  \cmark &
  28 &
  2009-2017 \\
\cite{survey4} &
  2019 &
   &
   &
  \cmark &
   &
   &
   &
   &
  \cmark &
  14 &
  2005-2019 \\
\cite{survey5} &
  2020 &
   &
   &
  \cmark &
   &
   &
   &
   &
  \cmark &
  5 &
  2005-2017 \\
\cite{survey7} &
  2020 &
  \cmark &
  \cmark &
  \cmark &
   &
  \cmark &
  \cmark &
  \cmark &
   &
   &
   \\
Ours &
  2021 &
  \cmark &
  \cmark &
  \cmark &
  \cmark &
  \cmark &
  \cmark &
  \cmark &
  \cmark &
  34 &
  2015-2021 \\ \hline
\end{tabular}
\endgroup
\label{table:surveys}
\end{table*}

\par Table \ref{table:surveys} presents a summary of the main contributions made by the aforementioned surveys in comparison to the ones made by our work. Unlike \cite{survey7}, and in addition to the contributions made by \cite{survey1, survey2}, we focus our efforts in compiling the most recent openly available datasets intended to foster the development in vehicle perception. Differently from \cite{survey1}, we include datasets based on radar and simulated data, as we consider it is of great importance for perception in the context of autonomous vehicles, as they allow development and testing without the expenses and risks of inherent to real-world scenarios. Additionally, unlike \cite{survey3, survey4, survey5}, we precisely define our scope to the vehicle navigation context, which lets us make in-depth discussions on the usage of multiple sensing modalities in order to overcome operation in adverse conditions.
\par Unlike other works, we do not focus on precisely describing the methods used in perception. We have chosen to briefly introduce the baseline models and cite the most recent ones for further consulting of the interested reader. Differently from all the aforementioned surveys, we describe and summarize the metrics used in object detection and image segmentation, grouped by dataset. To the best of our knowledge, this was never made in previous works, although being of utmost importance for introducing new researchers to the field.
\par It is worth mention that the aspects ''Sensing Modalities Description'', ''Data Diversity'', ''Methods'', and ''Metrics'' were considered as checked for the works that spent considerable effort on discussing them. For those who just briefly cited some of these aspects, they were considered unchecked.
\par For the ''Sensing Modalities Covered'' characteristic, we considered both the datasets and methods described in the survey; that is, if either the datasets or the methods related to ''Camera (2D)'' perception, it was considered as checked - a similar procedure was adopted for the ''Stereo Vision'' and ''Radar'' modalities. On the other hand, the ''Sensor Fusion'' modality was considered as checked only if a consistent effort has been made in describing concepts and characteristics related to sensor fusion.

\section{Advanced Driver Assistance Systems}
Every day, thousands of new vehicles are produced and start to integrate the urban environments and roads, where they will interact with other cars, pedestrians, cyclists, and a wide sort of other urban agents and elements, in very complex and dynamic scenarios.
\par In this context, traffic safety is of utmost importance to avoid any harm to people and their material goods. One of the main approaches towards improving road safety relies on improving vehicle safety.
\par Although widely addressed nowadays, vehicle safety has been studied since at least the half of last century.
\par The relatively recent history of assistance systems in cars can be divided into the Five Eras of Safety \cite{nhtsa_five_eras}.
\par In the first era, which covers the years from 1950 up to 2000, the main concern was the development of safety and convenience features, as Cruise Control (CC) and Anti-lock Brakes Systems (ABS). These former safety systems can be defined as Driver Assistance Systems (DAS), and they use information from the internal state of the vehicle, captured from sensors like Inertial Measurement Units (IMU) and odometers.
\par The second era was intended to the development of Advanced Safety Systems, also called Advanced Driver Assistance Systems (ADAS). The main difference between ADAS and DAS relies on the sensor types each set of methods use. While DAS rely on internal state sensors, ADAS exploit advanced sensors designed to perceive the environment, like cameras, radar, LIDAR, as well as map databases \cite{history_and_future_of_adas}. In this era, studies focused on features like the Electronic Stability Control, Blind Spot Detection, Forward Collision Warning and Lane Departure Warning.
\par The third era of safety introduced more contributions to the set of ADAS. Among the main technologies introduced are the Rear-view Video Systems, Automatic Emergency Braking (AEB), Pedestrian Automatic Emergency Braking (PAEB), Rear Automatic Emergency Braking (RAEB), Rear Cross Traffic Alert (RCTA) and Lane Centering Assist (LCA). When compared to the second era, one important characteristic introduced by the third era was the fact that systems like the AEB and PAEB, RAEB and LCA can have punctual actuation - over the the brakes or the steering - in order to prevent from accidents, instead of just alerting the driver of safety risks.

\begin{figure*}
  \includegraphics[width=\textwidth]{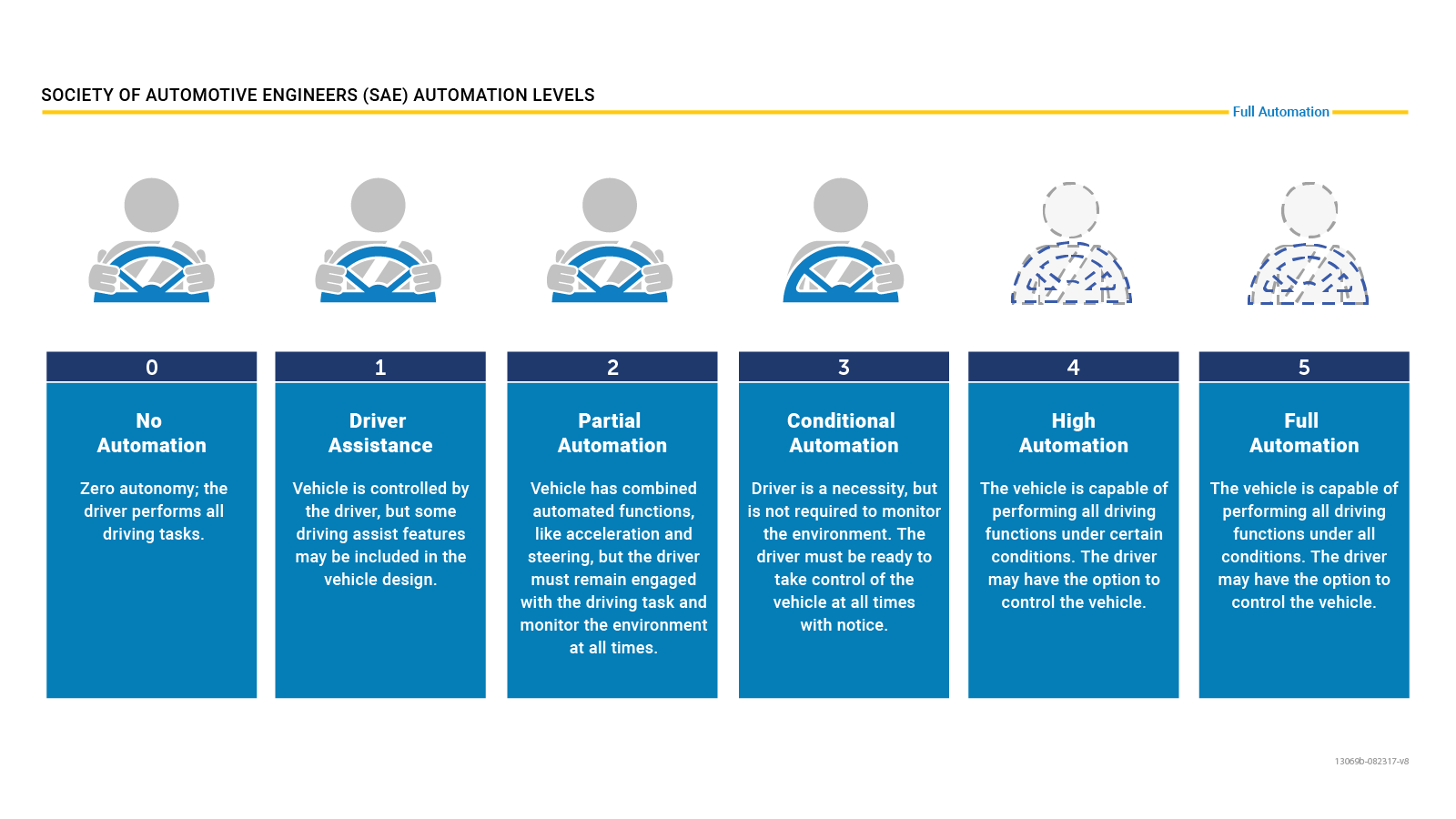}
  \caption{Vehicle automation levels \cite{nhtsa_five_eras}}
  \label{fig:automation_levels}
\end{figure*}

\par In the fourth era, which we are currently witnessing, the efforts are focused on, besides enhancing previous solutions, developing the so-called Partially Automated Safety Features, which cover Lane Keeping Assist, Adaptive Cruise Control, Traffic Jam Assist and Self-Park systems. One important improvement with respect to previous ADAS is the combination of multiple automated functions at the same time. For instance, in the third era, ADAS actuation was performed over either braking or steering, but not both. Conversely, ADAS from the fourth era combine acceleration and steering.

\par Finally, the fifth era will be devoted to the development of Fully Automated Safety Systems, represented by the Highway Autopilot. The main distinctive feature of these systems is their automation level; they are expected to allow the vehicle navigation without the need of the driver's attention, at least in most of the time.

\section{Autonomous Vehicles}
Although the wide variety of standard safety technologies embedded in today's vehicles, as previously described there is still a considerably long path to full vehicle automation. To achieve full autonomy, vehicles must progress through six levels of driver assistance technology, as defined by the Society of Automotive Engineers (SAE) \cite{sae} - figure \ref{fig:automation_levels}.
\par The automation levels directly relate to the characteristics presented in each of the Safety Eras previously described, and they can be distinguished according to the level of driver intervention.

\par In the first level, the driver is totally responsible for driving the vehicle, which has zero autonomy.
\par In the second level, some assistance features, like the Forward Collision Warning and the Electronic Stability Control, are introduced in the vehicle. However, the driver is still totally responsible for the vehicle guidance. It is worth mentioning that, although the assistance features can actuate, they do so over a single vehicle function - for example, the Automatic Emergency Braking system (AEB) actuates only on the brakes.
\par In the third level, the assistance systems can actuate on multiple vehicle functions - acceleration and steering, for instance -, although the driver is still responsible for driving and monitoring the environment at all times.
\par In the fourth level, also known as Conditional Automation, the driver is not required to monitor the environment at all times, but he still must be ready for taking the control of the vehicle when necessary.
\par In the penultimate stage, it is achieved a high automation level. The vehicle is capable of autonomously perform all driving functions in particular scenarios. Despite that, the driver should be able to take the control of the vehicle when appropriate.
\par Finally, in the Full Automation level the vehicle can autonomously navigate, regardless the scenario and driving conditions. It is important to mention that, even in this level of full autonomy, the driver must be able to take control of the vehicle, when judged necessary.
\par Autonomous Vehicles correspond to this last level of automation, as shown in figure \ref{fig:automation_levels}. They are basically composed of a perception, planning and actuation modules - in a very simplified way.
\par Through series of specialized internal and external sensors, the perception module can reconstruct its internal state and sense the environment, then feeding the decision module, which, using specialized methods and algorithms, generates actuation commands to be executed by the actuators constituting the actuation module.
\par Through this simplified pipeline, the vehicle can perceive possible risks. Then, given its internal and external context, it can decide the best actions in order to prevent from them.
\par Autonomous vehicles can also can be classified as critical systems, in which minor flaws can result in severe consequences both in terms of material goods or even in terms of people safety \cite{uber_accident}.
\par Therefore, it is essential that such systems are able to operate in all sorts of weather, traffic and lighting conditions, respecting other vehicles, vulnerable road users - pedestrians, cyclists - and transit rules.
\par As the main contributions of Autonomous Vehicles, it can be mentioned the increase in efficiency and road safety, economic and societal benefits, and convenience and mobility improvement \cite{nhtsa_five_eras}. 
\par The increase in safety can be easily understood by considering that automated driving removes the human factor from the crash equation. The economic and societal factors are related to the reduction in traffic victims who, besides representing losses to his companies in terms of labor, are prone to have a considerable decrease in quality of life due to injuries. When considering the efficiency and convenience, roads filled with autonomous vehicles cooperate to smooth traffic flow and reduce traffic congestion \cite{nhtsa_five_eras}, what, in turn, reduces the money and time spent in traffic. Lastly, the mobility improvement is related to offering new mobility options to people with disabilities or the elderly, improving their independence and range of opportunities.
\par Despite their many advantages, we still lack fully autonomous vehicles, either because of technical limitations or legal issues. A promising path to fully autonomy seems to address multi-modal sensor fusion, also known as hybrid sensor fusion, in order to achieve a more robust perception of the environment. When considering the legal aspects, although there are attempts to define a common legislation about self-driving cars, mainly in developed countries were the testing has been taking place for a longer time, this is still an open question.
% Developments are to be made also in planning and actuation, with studies being carried out in, for instance, more precise dynamic models for vehicles.
\par Although with a long way to go, there are a lot of autonomous vehicle projects all over the world, both from private and public institutions. Therefore, given the current collective effort towards self-driving cars, we can expect having these systems sharing the streets with conventional cars in a near future.

\section{Sensing Modalities}
\label{percep}
Systems that interact with the environment, such as autonomous vehicles, must have a robust perception of their surroundings.
\par
\begin{figure}
\centering
\includegraphics[width=0.47\textwidth]{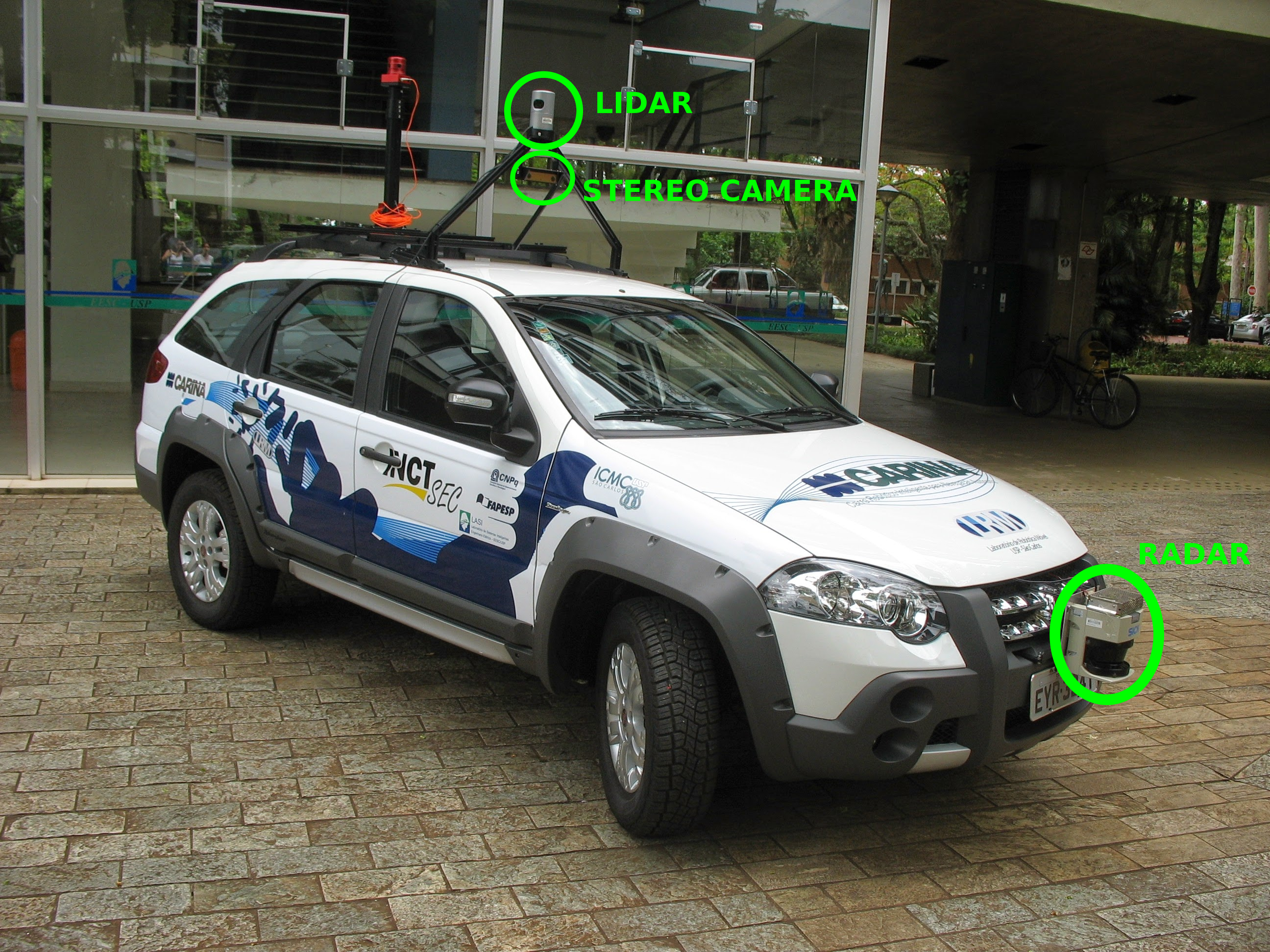}
\caption{Example of sensors commonly used in autonomous vehicles. Adapted from \cite{b41}.}
\label{fig:car_sensing}
\end{figure}

 Hence, it is usual to embed a wide range of sensors in such devices, each one with its particularities - figure \ref{fig:car_sensing}.

\par In the following sections, we describe camera and radar's main operating characteristics, advantages, and limitations. Additionally, we present the data fusion technique as a way to improve single-modality perception.
\par
\begin{figure}[h!]
\centering
\includegraphics[width=\columnwidth]{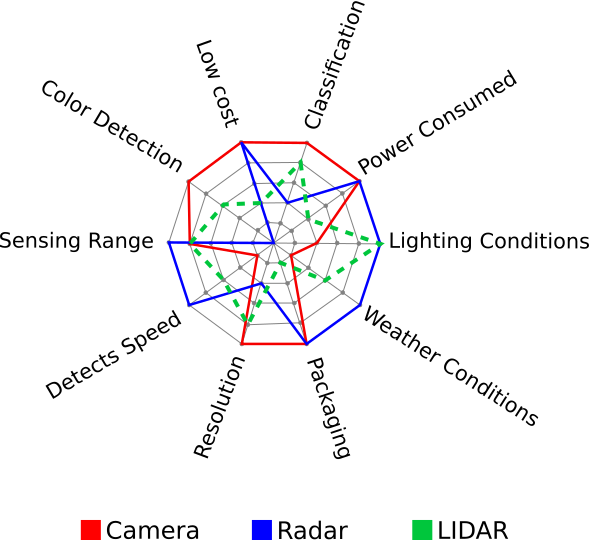}
\caption{Comparison of Camera, Radar and LIDAR sensors, with respect to various characteristics of operation. Adapted from \cite{survey7}}.
\label{fig:sensing_mod_comparison}
\end{figure}

 Figure \ref{fig:sensing_mod_comparison} shows a comparison among the Camera, Radar and LIDAR sensors, considering various characteristics of their operation.

\par It is worth mention that, although many studies have been carried out on LIDAR perception, we focus our analysis in camera and radar perception due to their considerably lower cost in comparison with the LIDAR sensor. Interestingly, some authors consider that radar data can substitute LIDAR data in vehicle perception \cite{b10}. Indeed, as shown in Figure \ref{fig:sensing_mod_comparison}
, camera and radar, if combined, meet all the characteristics considered, thus being a viable alternative to LIDAR-based perception.

\subsection{Camera}
Cameras are passive sensors, which operate by receiving lighting information from the environment. This characteristic poses cameras as appropriate sensors when the goal is to capture color, shape and texture information from the surroundings. On the other hand, it also implies sensitive limitations related to degradation caused by adverse lighting or weather conditions. 
\par
\begin{figure*}
\centering
\includegraphics[width=\textwidth]{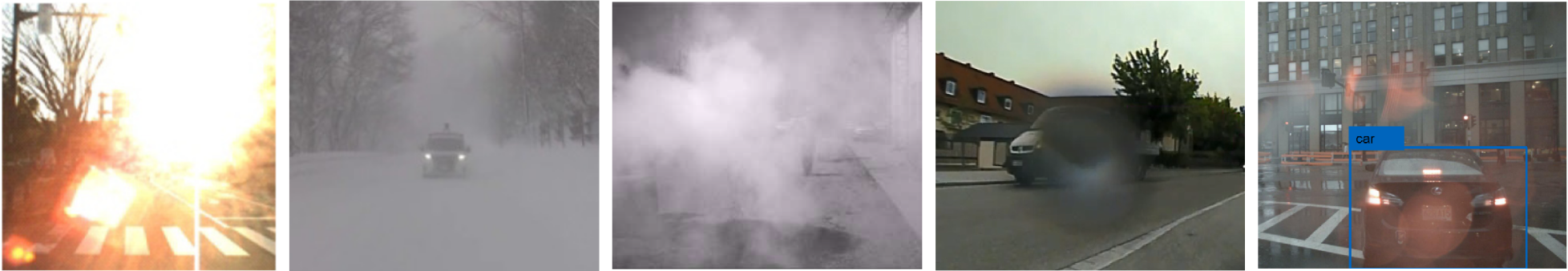}
\caption{Examples of operation in adverse conditions. Adapted from: \cite{b12, b37, sun_glare}.}
\label{fig:lim_camera}
\end{figure*}

 As depicted figure \ref{fig:lim_camera}, the sun glare or low lighting conditions - at sun rise or night fall, respectively -, can be prejudicial to camera perception. Additionally, cameras suffer from occlusion caused by dust, rain, snowfall or fog.

\subsubsection{Monocular Vision}
Monocular vision is the most common sensing modality in Computer Vision, since it fostered important advancements throughout the years, before the adoption of other types of data, like 3D or multi-spectral imagery.
\par
\begin{figure}[h!]
\centering
\includegraphics[width=\columnwidth]{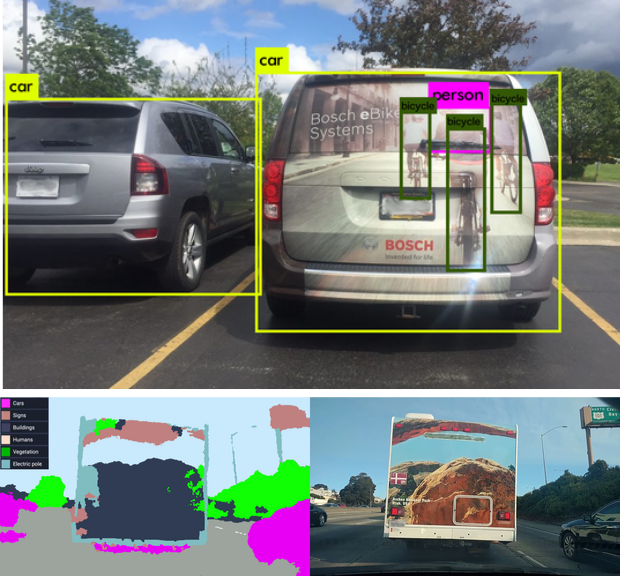}
\caption{Examples of errors in 2D perception \cite{the_guardian}.}
\label{fig:error_2D_perception}
\end{figure}

 The main limitation of monocular-based methods, however, is the absence of the notion of depth in the data. This can particularly compromise the perception in urban environments, where there are lots of visual information in the form of advertisements. Without wondering about the depth of and object, it may wrongly classify 2D plots as actual entities - figure \ref{fig:error_2D_perception}.

\subsubsection{Stereo Vision}
\begin{figure}[h!]
\centering
\includegraphics[width=\columnwidth]{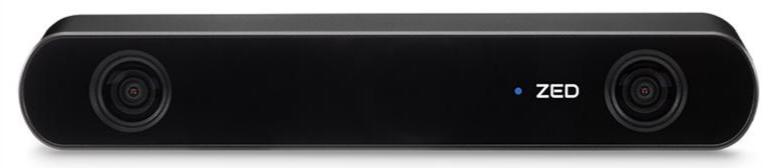}
\caption{Stereo Camera ZED2, from StereoLabs.}
\label{fig:stereo}
\end{figure}

Stereo cameras, such as the one depicted in figure \ref{fig:stereo}, supply the absence of depth in monocular vision, providing 3D perception of the environment.

\par
\begin{figure}[h!]
\centering
\includegraphics[width=\columnwidth]{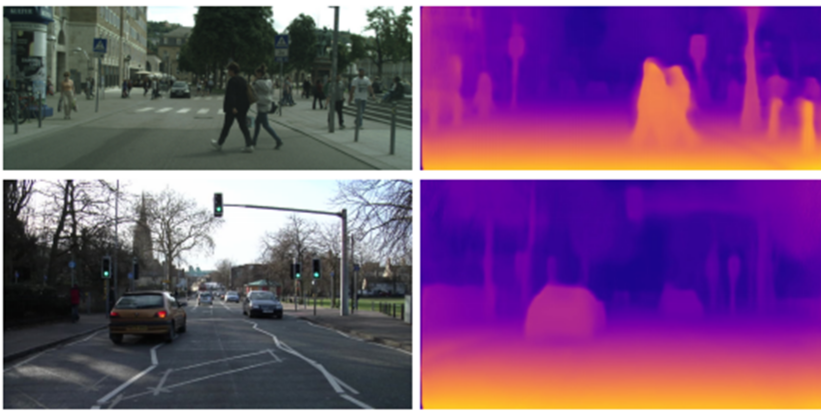}
\caption{Scene and its depth map \cite{b35}.}
\label{fig:depth_map}
\end{figure}

 This type of sensor operates by capturing images from monocular cameras, offset by a certain amount. From this pair of images, and the intrinsic parameters of the cameras, a disparity map is generated, which is then converted into a scene depth map (figure \ref{fig:depth_map}).
\par It should be noted that the process of generating depth maps can be done through Deep Learning-based techniques that can extract the corresponding depth map from a single image, as shown in \cite{b35}.

\par Their main advantage consists in generating images that aggregate both contours and depth, factors that motivate its application for a wide range of purposes. However, it also has severe limitations.
\par The first limitation concerns its low range, from 20 to 30 meters. It is particularly a drawback in terms of autonomous vehicle applications, where a greater range is required so that the vehicle can act in time to avoid further risks. The second limitation relates to the absence of color or texture cues in the data. The last limitation is the degradation suffered under adverse conditions, as already mentioned.

\subsection{Radar}
Radar sensors operate by emitting and receiving electromagnetic pulses, following principles similar to sound wave reflection. Initially, a transmitter generates radio frequency pulses with high power, which are transmitted to the medium (commonly the air) through an antenna. Upon reaching an object, such pulses generate a return pulse (or echo) resulting from the transmission of radio frequency energy to this object. A small portion of the reflected energy returns to the radar through the antenna and is directed to the receiver. Finally, the receiver sends the energy to the signal processor to determine the direction, distance, and even speed of the object identified.
\par Its main advantages are its long range - hundreds of meters - and robustness to weather and lighting conditions. These features permit to determine the position of obstacles invisible to the naked eye - or even to other sensors like cameras - due to distance, darkness, or weather \cite{b36}.
\par In recent years, Deep Learning-based object detection using radar data has received increasing attention. \cite{radar_1} proposes a network called DANet, which, by the extraction of temporal and multi-scale spatial features, detects objects in range-angle radar images. \cite{radar_3} applies Recurrent Neural Networks in the processing of Ultra-Wide Band radar signals for road obstacle detection. \cite{radar_6} introduce a Radar-based real-time region proposal method, which can be integrated into any object detection network, such as Fast R-CNN. \cite{radar_7} studies the application of Faster R-CNN \cite{faster_r-cnn} and SSD \cite{ssd} to the processing of radar imagery for object detection. \cite{radar_8} proposes an architecture for vehicle detection based on the processing of Range-Azimuth-Doppler Tensors.
\par The literature on radar-based segmentation, however, is still scarce. The work from \cite{radar_2} proposes the RadarPCNN model, based on the PointNet++ \cite{pointnet++}, to perform semantic segmentation on radar point clouds. \cite{radar_5} addresses the problem of open space segmentation for robot navigation, with focus in low-memory footprint and real-time processing. 
\par
\begin{figure*}
\centering
\includegraphics[width=\textwidth]{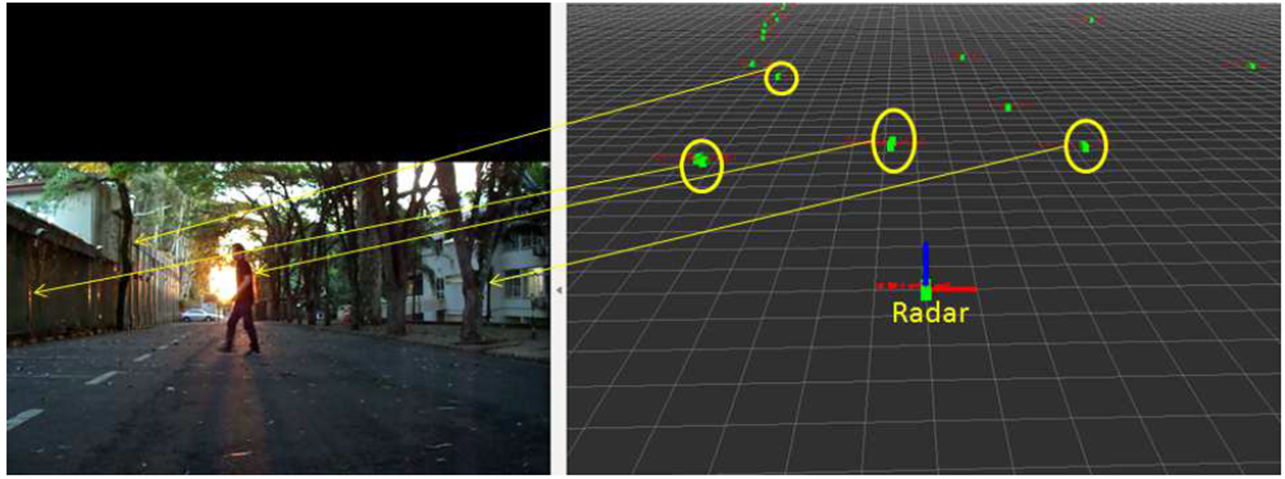}
\caption{Radar detection represented as points \cite{b37}.}
\label{fig:lim_radar}
\end{figure*}

 The main limitation of this type of sensor lies on the impossibility of determining the shape of the detected objects. Figure \ref{fig:lim_radar} shows an example of radar readings that represent the detected objects as dots.

\subsection{Sensor Fusion}
In order to mitigate the limitations and benefit from the advantages offered by camera and radar sensors, methods for hybrid data fusion have currently been proposed. Such methods aim to aggregate both sensing modalities, so as to generate richer representations of the environment, ultimately contributing to a more robust perception.
\par
\begin{figure}[h!]
\centering
\includegraphics[width=\columnwidth]{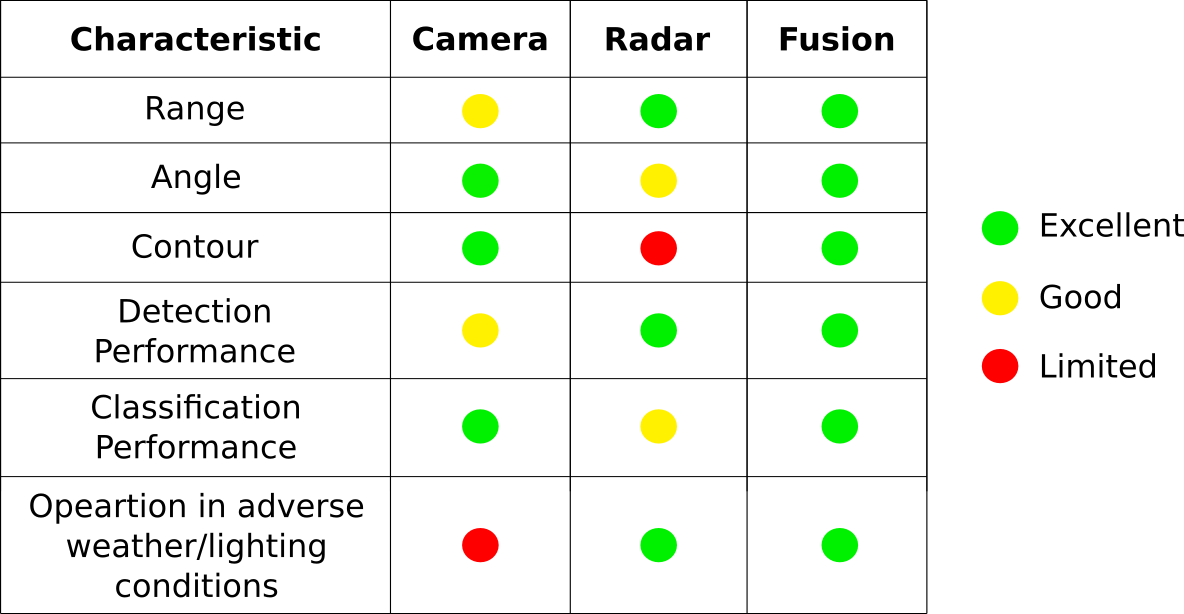}
\caption{Advantages of sensor fusion compared to perception based on single sensing modality. Adapted from \cite{b14}.}
\label{fig:tab_fus}
\end{figure}

 In \cite{b14}, data fusion is presented as one of the central pillars for future developments linked to autonomous vehicles. The authors also discuss the advantages of camera-radar fusion compared to single-modality perception (figure \ref{fig:tab_fus}).

\par Data fusion has been handled using Computer Graphics and Computer Vision methods for a long time. In \cite{b37} are proposed techniques for calibration and fusion of Radar, LIDAR, and Stereo Camera sensors. The authors demonstrate the effectiveness of the method against single sensor-based approaches, resulting in the reduction of the false-positive and false-negative rates - fundamental metrics in object detection.
\par Recently, though, several works have been proposing Deep Learning-based fusion techniques. According to \cite{survey2, survey7}, in this scenario, there are basically three important questions to be addressed : ''What to fuse?'', ''How to fuse?'', and ''When to fuse?''.

\subsubsection{''What to fuse?''}
This question focuses on answering what sensing modalities should be fused. It also covers how to represent them in an appropriate way so that fusion can be performed correctly.
\par Camera data are usually represented as plain RGB images. However, in order to obtain richer representations, some works generate additional sensing information, such as optical flow \cite{opt_flow_1, opt_flow_2, opt_flow_3, opt_flow_4, opt_flow_5}. Other works combine both RGB and Depth information \cite{b7, multimodal_rgbd, adapnet}.
\par Additionally, as a means to achieve a more robust perception in challenging conditions, infrared images from thermal cameras can be employed. In this sense, multi-spectral approaches are receiving increasing attention \cite{upnet, multispec1, MODT, MSSD}.
\par Radar data can encode information from the environment in the form of amplitudes, ranges and the Doppler spectrum \cite{survey2}. Its data can be represented as 2D maps and processed by Convolution Neural Networks for object detection \cite{radar_7, radar_8}, segmentation \cite{radar_5}, and classification \cite{b11}. Alternatively, radar data can also be represented as point clouds \cite{camera_radar_1, radar_2}.
\par Most of the works on deep fusion - Deep Learning-based fusion - consider LIDAR and camera fusion. In fact, many of the best performing works in popular benchmarks are based in some kind of LIDAR-camera fusion.
\par LIDAR-radar fusion, however, is not considered by some works as a valid combination for vehicle perception, once this approach presents limitations in many critical aspects, such as resolution and color detection \cite{survey7}.
\par Camera-radar fusion, though, can be used in different scenarios, being considered a solution with good potential for the vehicle perception problem \cite{survey7}. The camera offers rich visual information in the form of shapes, colors and textures, but suffers with degradation in adverse conditions. Radar, on the other hand, does not allow delineating the objects' shapes, but is robust to lighting and weather challenging conditions.
\par Because of its advantages, camera-radar fusion is receiving increasing attention in the literature. Many works already consider this multi-modal perception in detection tasks \cite{b12, b13, b14, b15, camera_radar_1, camera_radar_2, camera_radar_3, camera_radar_4, camera_radar_5, camera_radar_6, camera_radar_7, camera_radar_9}, while few of them apply the sensor fusion approach in segmentation tasks \cite{b16}.  Therefore, there is plenty of room for exploring camera-radar fusion in vehicular perception.

% \begin{figure}
% \centering
% \includegraphics[width=\columnwidth]{opt_flow}
% \caption{Optical Flow \cite{opt_flow_1}.}
% \label{fig:opt_flow}
% \end{figure}

\subsubsection{''How to fuse?''}
This question dwells on what operation to use when fusing the different sensing modalities.
\par The most used operations range from simple (i) point-wise addition (or average) and (ii) concatenation (stacking) of feature maps, to more complex (iii) ensembles and the termed Mixture of Experts (MoE) \cite{moe}.
\par The last operation weights over the informativeness of each sensing modality - which are processed by separate networks (experts) -, given the current context \cite{moe2}. For instance, in a camera-radar fusion method, RGB images will have less informative value than radar data when operating under adverse lighting or weather conditions.

\subsubsection{''When to fuse?''}
Neural Networks, and more specifically Deep Convolutional Neural Networks, represent and process features in a hierarchical manner throughout their different levels of layers. Initial layers process coarser representation of the input, thus having more detailed spatial information. As we move further in the architecture, the feature maps lose spatial detail to gain semantic information. Hence, in the last layers, the feature maps completely encapsulate semantics, but are limited in terms of spatial information - lack in terms of contour delineation, for example.
\par
\begin{figure*}
  \includegraphics[width=\textwidth]{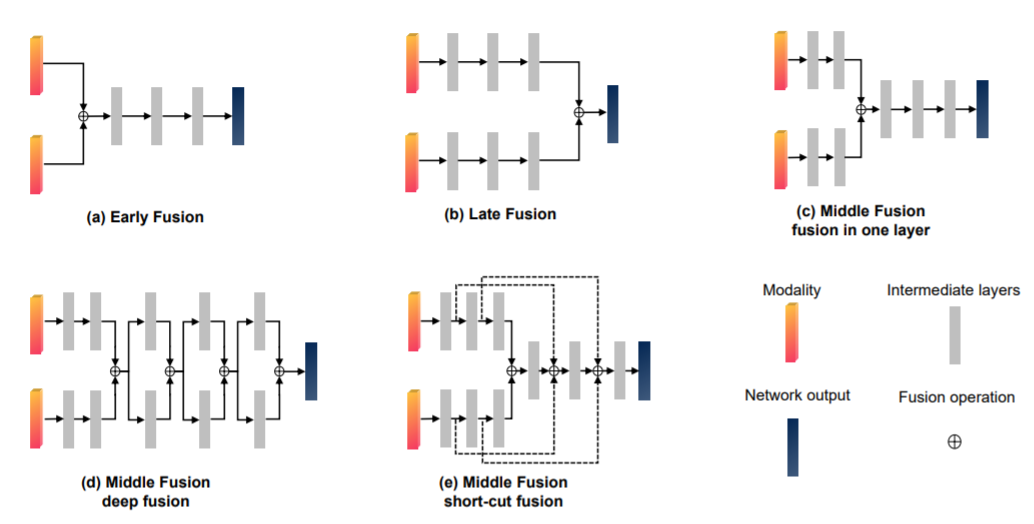}
  \caption{Schemes for early, middle and late fusion \cite{survey2}.}
  \label{fig:when_to_fuse}
\end{figure*}

 Given these characteristics, it is a common practice to fuse features from different levels of the network, in order to make good use of both levels of representation. When we add more than one sensing modality in the equation, feature fusion becomes even more powerful.
Choosing at which level of feature representation should the fusion take place is addressed by the question ''when to fuse?'' - Figure \ref{fig:when_to_fuse}.
\par Early Fusion, or data-level fusion, fuses the input data from the different sensing modalities. Alternatively, it can also fuse features from the initial layers of a network. The main pros of early fusion are the full exploration of raw data and low computation cost, since the network jointly processes the fused sensing modalities, thus sharing the network computation. However, it also has drawbacks, the first being model inflexibility - need for retraining in case of sensing modality replacement. The second disadvantage is sensitiveness to spatial-temporal misalignment due to calibration errors, sensing rate or sensor failure \cite{survey2}.
\par Middle Fusion, or feature-level fusion, involves fusing features from intermediate layers of the network. It can be one of: one-layer fusion, deep fusion or sort-cut fusion - figure \ref{fig:when_to_fuse} (c), (d) and (e), respectively. The main drawback of middle fusion is the difficulty in finding the optimal fusion scheme for each particular network architecture.
\par Late Fusion, or decision-level fusion, occurs in a later step of the network processing pipeline, closer to the output. It combines the outputs of domain-specific networks (experts) for the different sensing modalities. Its main advantages relies on model flexibility, given that when a new sensing modality is introduced, only its expert network must be retrained. On the other hand, the main drawbacks are the high costs in terms of computation and memory, as well as the discarding of possibly important features from intermediate layers.
\par In \cite{b12}, the authors address the question of ''when to fuse'' by proposing a Deep Learning model for data fusion that automatically learns at which level of the network structure the fusion is most beneficial.

\section{Perception Tasks}
The data obtained through the sensors described in section \ref{percep} are used in several perception tasks in the context of Autonomous Vehicles and ADAS. Particularly, when considering the field of Computer Vision, currently boosted by Deep Learning, tasks like object detection and image segmentation are of paramount importance.

\subsection{Detection}
\label{detection}
Object detection seeks to locate, either with 2D or 3D bounding boxes, and classify the elements of an image.
\par
\begin{figure*}
  \includegraphics[width=\textwidth]{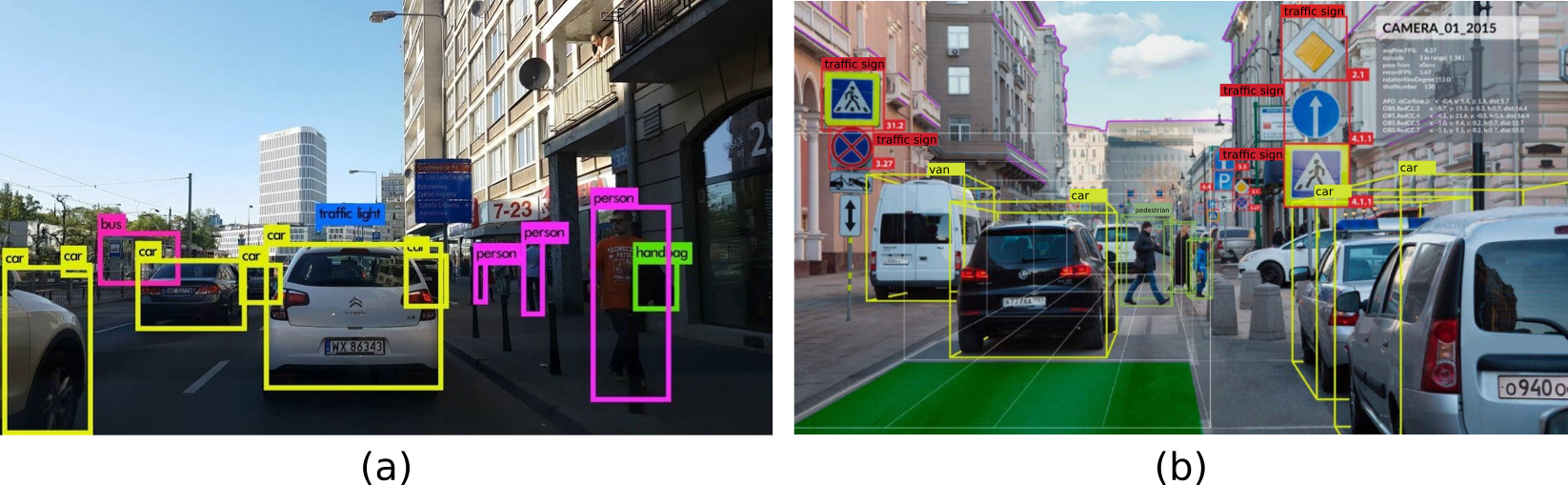}
  \caption{Examples of 2D (a) and 3D (b) detection.}
  \label{fig:tasks}
\end{figure*}

 In the first case, each element is associated with a label identifying its category and a 2D bounding box, representing its location (figure \ref{fig:tasks} (a)).
\par In 3D detection, in addition to labels and 3D bounding boxes (figure \ref{fig:tasks} (b)), each entity is associated with information such as velocity, orientation, or even action descriptors - in the case of a vehicle, whether it is moving, parking or stopped.
\par Detection is a task widely studied in the literature and can be considered solved in environments with good navigation conditions, since there are already Deep Learning models that perform extremely fast and accurate in such situations.
\par Deep Learning-based object detectors can be divided into two main groups: one-stage and two-stage detectors.
\par Two-stage detectors were the first set of Deep Learning-based detectors proposed in the literature. They try to solve the task of multi object detection in two steps. The first involves the generation of region proposals, which represent areas in the image with high probability of having objects. Then, in a second step, these region proposals are processed by a CNN in order to obtain both the object location and classification.
\par Inside this group, the Region Proposal family of architectures - termed R-CNN - stands out. R-CNN \cite{r-cnn} was the pioneer method. Its successors tried to solve problems of previous versions. Faster R-CNN \cite{fast_r-cnn} allowed end-to-end training. Faster R-CNN \cite{faster_r-cnn} integrated the region proposal generation to the full pipeline. Mask R-CNN \cite{mask_r-cnn}, although proposed for instance segmentation purposes, also has branches for classification and detection and achieved better performance in detection accuracy.
\par One-stage detectors try to solve the problem of multi object detection in a single passage, providing faster performance - faster inference, higher Frames Per Second (FPS). YOLO \cite{yolo}, its posterior variants \cite{yolov2, yolov3, yolov4}, and the Single Shot Detector \cite{ssd} are some examples of famous and widely used one-stage detectors.
\par Despite achieving real-time performance (inference rate above 30 FPS), one-stage detectors deliver a considerable lower accuracy (Mean Average Precision, mAP). Thus, finding a good trade-off between accuracy and performance is still an open challenge.
\par Additionally, under adverse operation - lighting or weather challenging conditions - there can be a significant drop in performance. Thus, detection in adverse navigation conditions remains a field with several research opportunities.

\begin{figure}
  \includegraphics[width=\columnwidth]{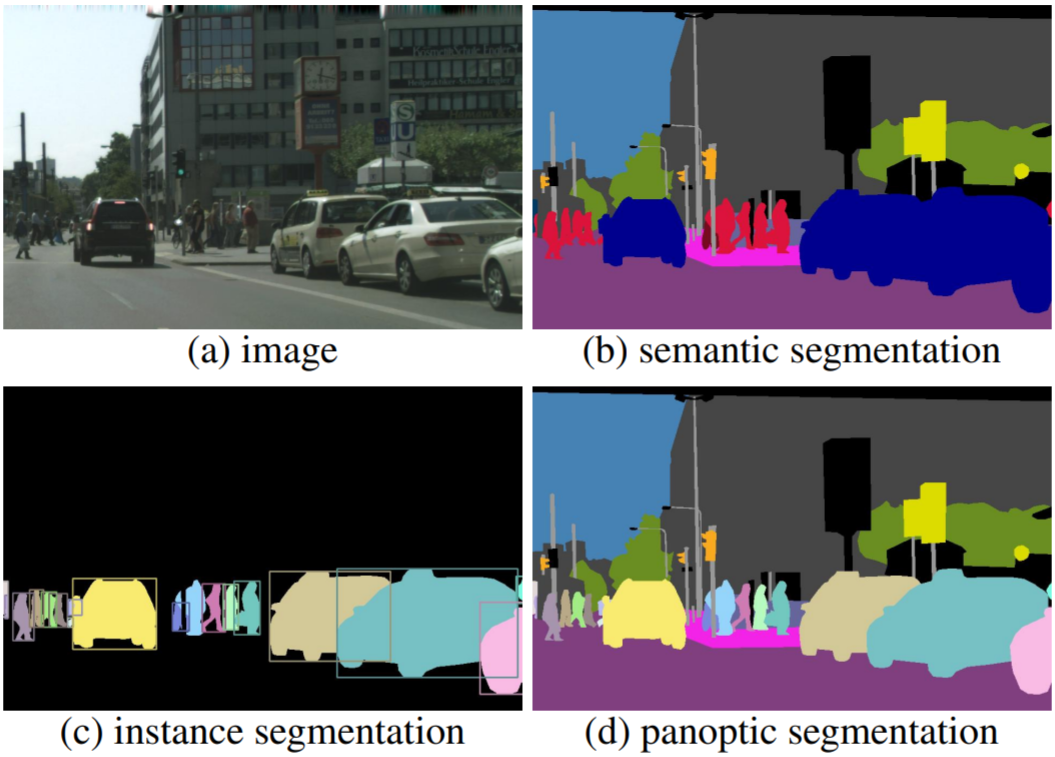}
  \caption{Examples of semantic (a), instance (b) and panoptic (c) segmentation \cite{panoptic_seg}.}
  \label{fig:tasks_2}
\end{figure}

\subsection{Image Segmentation}
\label{segmentation}
Image segmentation is the task of classifying the image at a pixel level. Each pixel is associated with a label, referring to one of the classes being considered - figure \ref{fig:tasks} (c).
\par There are basically three types of segmentation: semantic segmentation, instance segmentation and panoptic segmentation. All three types are described below.

\subsubsection{Semantic Segmentation}
\begin{figure}
  \includegraphics[width=\columnwidth]{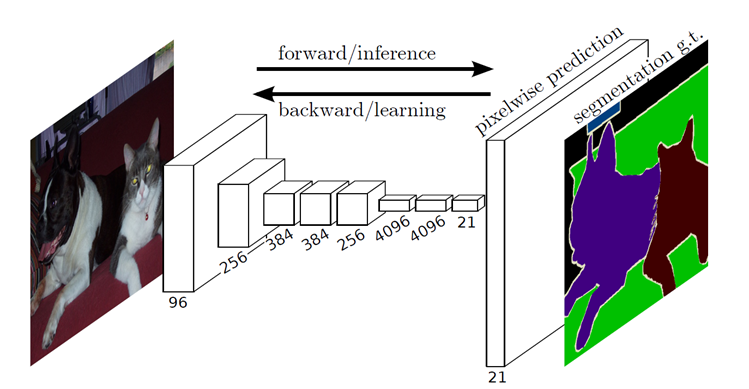}
  \caption{FCN basic architecture \cite{fcn}.}
  \label{fig:fcn}
\end{figure}

Semantic segmentation consists of pixel-level image classification without any distinction between instances of the same class.
The first remarkable work in this type of image segmentation was published back in 2015. The Fully Convolution Networks (FCN) \cite{fcn} proposed to change fully connected layers by its convolutional equivalents, through a process named by the authors as ''convolutionalization''. In this way, after removing all the dense layers, the network could process images of any size and, instead of outputting a vector of classes, it generated a dense classification (pixel-wise prediction) in the form of a ''heatmap''. In this type of representation, each 2D position represented a pixel in the original image, and the probabilities associated with each class were stored along its depth. Figure \ref{fig:fcn} depicts the basic architecture of FCN. 
\par
\begin{figure*}
  \includegraphics[width=\textwidth]{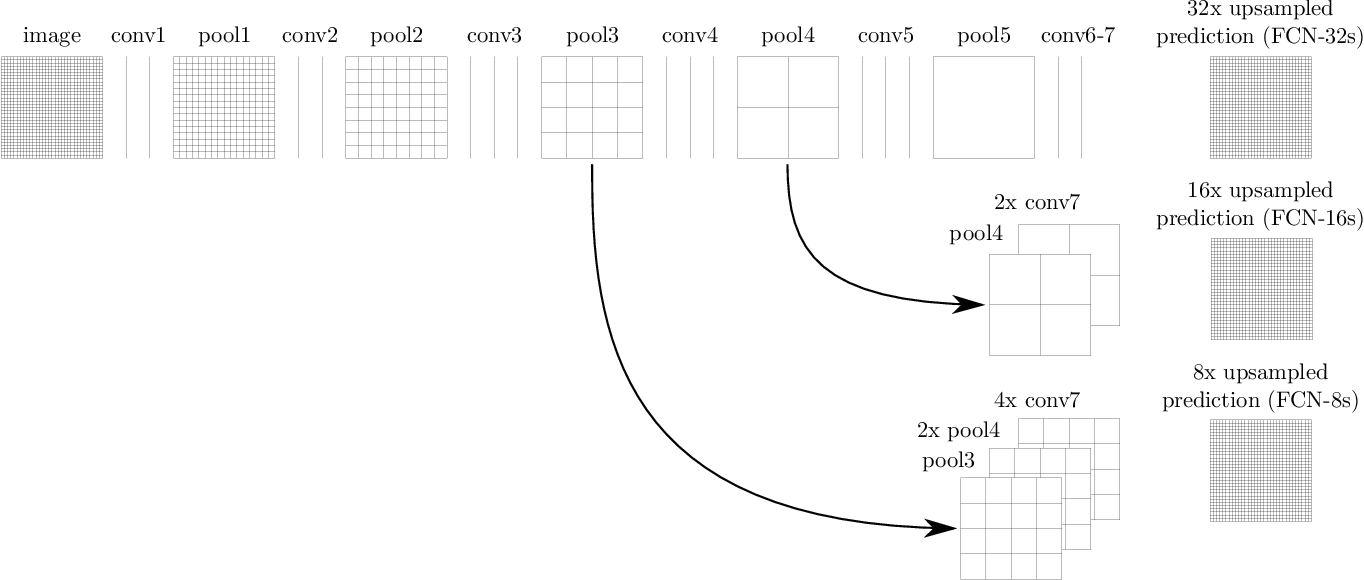}
  \caption{The three FCN architectures, based of the upsampling factor. \cite{fcn}.}
  \label{fig:fcn_types}
\end{figure*}

 In addition, the authors proposed three types of architecture, which differ according to the way the summarized feature representations were upsampled to the original image size. The resulting architectures, called FCN 8s, FCN 16s and FCN 32s (Figure \ref{fig:fcn_types}), combined features from different depth levels in the network and then upsampled them using different strides - 8, 16 and 32, respectively.

\par A contemporaneous work to FCN was U-Net \cite{u-net}. Originally proposed for Biomedical Image Segmentation, the architecture consisted of two basic paths: a contracting (downsampling) path and a symmetric expanding (upsampling) path.
\par
\begin{figure}
  \includegraphics[width=\columnwidth]{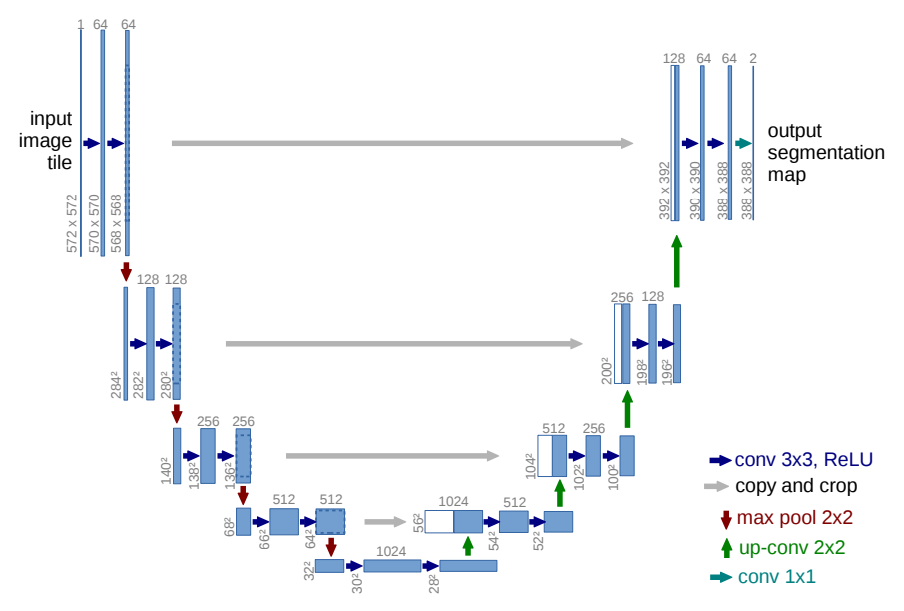}
  \caption{U-Net architecture \cite{u-net}.}
  \label{fig:u-net}
\end{figure}

 A distinctive characteristic of this architecture consists in what the authors called ''copy and crop''. Through this operation, feature maps of the downsampling path are cropped - to match the dimensions of the feature maps in the expanding path -, copied and concatenated with the correspondent feature maps in the expanding path (Figure \ref{fig:u-net}). This mechanism allows to aggregate spatial (from initial layers) and semantic (from final layers) information in order to obtain more precise segmentation masks.

\par Motivated by the success of previous models, many works based on Convolutional Neural Networks were proposed in the subsequent years.
\par
\begin{figure*}
  \includegraphics[width=\textwidth]{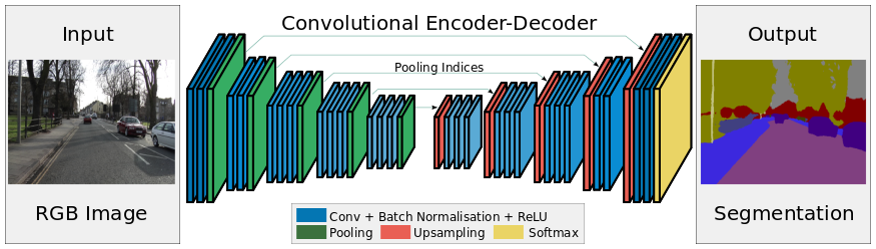}
  \caption{SegNet architecture \cite{segnet}.}
  \label{fig:segnet}
\end{figure*}

 SegNet \cite{segnet} was another remarkable work. Proposed for scene understanding applications, the network consists of an encoder-decoder architecture. Its main innovation lies on the upsampling mechanism. The authors introduce connections termed ''Pooling Indices''. During the max-pooling operation in the encoder, the indices of the max terms are stored and, through the Pooling Indices, transmitted to the corresponding layers of the decoder. This eliminates the need to learn how to upsample, and generates a sparse feature map with just the location of the pooling indices populated. Afterwards, a convolutional learnable filter is applied to this sparse map in order to produce dense feature maps. The architecture of SegNet is show in figure \ref{fig:segnet}.

\subsubsection{Instance Segmentation}
Instance segmentation improves upon semantic segmentation, as it distinguishes between elements of countable classes, like cars, pedestrians and cyclists. However, it does not classify elements from amorphous or uncountable classes, such as the sky, the buildings and the street.
\par One of the most prominent works in instance segmentation is Mask R-CNN \cite{mask_r-cnn}. The method proposes an approach to simultaneous detection and instance segmentation, extending Faster R-CNN by adding a mask prediction branch to the existing bounding box branch, with little overhead.

\subsubsection{Panoptic Segmentation}
Panoptic segmentation \cite{panoptic_seg} puts together the best of both previous segmentation modalities, generating masks (segmentations) for countable elements (instance segmentation), such as cars and pedestrians, and amorphous/non-countable elements (semantic segmentation).
\par Xiong et al. \cite{upsnet} proposes a unified panoptic segmentation network UPSNet, consisting of a residual backbone, a semantic segmentation head, and a Mask R-CNN-based instance segmentation head. The outputs from both heads then feed a panoptic segmentation head, which performs pixel-wise classification and solves possible inconsistencies between the previous heads.

\subsubsection{Real-time Segmentation}
In vehicular applications, real-time operation is a crucial requirement, since it allows quick reaction to possible risks.
\par Although being a research subject with high level of maturity, great part of the efforts in image segmentation focused in increasing accuracy. When we add the real-time constraint in the equation, a new need for lightweight, efficient and fast image segmentation methods and architectures emerge.
\par Several contributions have already been made to real-time semantic segmentation \cite{icnet, bisenet, qnet}. Despite of that, there is still many opportunities for improvement.
\par In the branch of instance segmentation, another relevant work in the field is YOLACT \cite{yolact}. Derived from one-stage detectors, it is faster than previous works, but with lower segmentation accuracy. An improvement over YOLACT is YOLACT++ \cite{yolact++}.
\par Finally, because panoptic segmentation was the latest image segmentation task proposed, the literature still lack works on real-time panoptic segmentation \cite{panoptic_real_time_1, panoptic_real_time_2, panoptic_real_time_3}.

\section{Datasets}
\label{datasets}
Until very recently, databases constructed for perceptual tasks in autonomous vehicles were mainly based on 2D image data - mostly captured under ideal conditions of navigation. However, the growing concern with navigation in challenging scenarios, and the consequent migration towards data fusion strategies, have encouraged the creation of datasets encompassing adverse navigation conditions and multiple sensing modalities.
\par The main databases used in vehicle perception are presented below. We conduct an analysis - summarized in Table \ref{table:1} - regarding the sensing modalities covered, the presence of adverse conditions, the types of applications intended, and the dataset size.
\par Additionally, the availability of camera and radar labeled data - essential to supervised learning in Deep Learning -, in the period from 2012 to 2021, is illustrated in figure \ref{fig:evol_mod}.

\subsection{KITTI Dataset}
One of the pioneering works in the context of autonomous vehicles, the KITTI dataset \cite{b17} provides a more than thirty thousand labeled images - 2D and 3D bounding boxes -, and 800 images with masks for segmentation, covering various lighting and traffic conditions.
\par It allows the development in various branches of vehicle-related perception tasks, including stereo vision, optical flow, scene flow, visual odometry, SLAM, object detection and tracking, road/track detection, and semantic segmentation.
\par Although first proposed in 2012 - thus not fitting in the range from 2015 to 2021, as intended by our work - it was, and still is, a work of utmost importance in the context of perception for autonomous vehicles, so that we considered it worth mentioning this dataset.

\subsection{Common Objects in Context (MS-COCO)}
MS-COCO \cite{ms_coco} is a large-scale dataset aimed at object detection, segmentation and labeling.
\par It accounts for around 330,000 images, with 200,000 of them labeled.
\par Although not specifically created for research on autonomous vehicles, it contains data with urban road scenes.

\subsection{KAIST multispectral pedestrian}
Focusing on the detection and tracking of vulnerable road users (pedestrians and cyclists), the KAIST Multispectral Pedestrian Dataset \cite{kaist_pedestrian} consists of 95,000 color-thermal image pairs captured under different lighting and traffic conditions - urban/downtown, road, and campus.
\par It contains a total of 103,128 dense annotations, including the temporal correspondence between bounding boxes in different frames.

\subsection{Udacity}
Proposed for didactic purposes, the Udacity dataset \cite{udacity} covers various lighting changes and is delivered in two versions. The first one contains around 9423 frames annotated with 2D bounding boxes (for cars, trucks and pedestrians), while the second one has 15000 frames, additional fields for occlusion, and labels for traffic lights.

\subsection{JAAD}
The Joint Attention in Autonomous Driving (JAAD) Dataset \cite{jaad} focuses on pedestrian and driver behaviors at the point of crossing and factors that influence them.
\par It covers scenes filmed in North America and Eastern Europe, under various weather conditions, accounting for a total of 82,032 frames.
\par For each frame, bounding box annotations, occlusion tags, behaviour labels, demographic attributes and a list of visible traffic elements are provided.

\subsection{Tsinghua-Daimler Cyclist Detection Benchmark Dataset}
The Tsinghua-Daimler Cyclist Benchmark is aimed at the specific task of cyclist detection.
\par It comprises almost 15,000 RGB images annotated with 2D bounding boxes. No mention to adverse operation conditions is made.

\subsection{Playing for Data: Ground Truth from Computer Games}
Comprising simulated data, this dataset was created through the annotation of frames extracted from the game GTA V.
\par Aimed at semantic segmentation of (simulated) urban scenes, it accounts for 25,000 labeled frames covering different lighting and weather conditions.

\subsection{SYNTHetic collection of Imagery and Annotations (SYNTHIA)}
The SYNTHIA \cite{b19} dataset was created to foster the development of semantic segmentation and scene comprehension solutions in the context of autonomous vehicles.
\par It comprises around 214,000 labeled images, according to 13 classes, from urban driving scenarios, roads, and vegetation areas. It also covers multiple seasons, weather, and lighting conditions.
\par Its most distinctive characteristic refers to the simulated nature of its data. In the context of autonomous vehicles, simulation is of utmost importance, because it allows the development and testing without the need for acquiring expensive sensors, also preventing from time-consuming and risky data capture procedures in the outdoors.

\subsection{Cityscapes}
The Cityscapes \cite{b18} dataset focuses on the task of urban scene understanding. In addition to masks for dense semantic segmentation, it provides masks for instance segmentation - people and vehicles.
\par Regarding data diversity, it comprises scenes from 50 cities, acquired at different hours of the day, seasons of the year, and weather conditions. In addition, it presents a large number of dynamic objects in all images.
\par In total, the database has 5,000 finely labeled images and 20,000 coarsely labeled images.
\par An interesting fact is that the database has extensions made by other collaborating researchers. There are versions labeled with bounding boxes, as well as versions with fog and rain artificially inserted.

\subsection{Multi-spectral Object Detection dataset}
Focused on 2D object detection, the Multispectral Object Detection Dataset \cite{MODT} accounts for around 7,500 images captured from RGB, near-infrared, middle-infrared, and far-infrared cameras, under different lighting conditions.

\subsection{Multi-spectral Semantic Segmentation dataset}
The Multi-spectral Semantic Segmentation dataset \cite{MSSD} was designed to improve the performance of semantic segmentation under adverse weather and lighting conditions, and accounts for around 1,500 RGB-Thermal image pairs. 

\subsection{Mapillary Vistas}
The Mapillary Vistas dataset \cite{mapillary} contains 25,000 high resolution images annotated with masks for instance and semantic image segmentation of road scenes.
\par Collected all around the world, it gathers data under various weather and lighting settings, standing as one of the richest datasets in coverage of real-world conditions.

\subsection{KAIST}
The KAIST dataset aims at all-day perception. To this end, it gathers multi-spectral images from different drivable regions (campus, urban and residential), and in different periods of the day, including specific time slots, such as sunrise, sunset, and dawn.
\par In total, it provides about 7.5 thousand frames.

\subsection{ApolloScape}
First proposed in 2018, the ApolloScape dataset \cite{apollo} has as objective to foster multi-sensor fusion and multi-task learning in the field of Computer Vision.
\par It  is divided into subsets accordingly to different tasks, which range from semantic segmentation to self localization.
\par It comprises a total of 150,000 images representing various illumination, weather and traffic conditions.

\subsection{nuScenes}
Inspired by the KITTI dataset, nuScenes \cite{b21} was the first dataset to provide data from the complete set of sensors used in autonomous vehicles.
\par In total, around 1.4 million camera images, 390,000 LIDAR scans, and 1.3 million radar scans are provided. From this data, 40,000 image-LIDAR-radar triplets are labeled with 3D bounding boxes. Additionally, there are labels for object-level attributes such as visibility, activity - vehicle/pedestrian moving/stopped - and pose.
\par According to the authors, the data capture was carried out on two continents to include a wide variety of locations, weather conditions, times, types of vehicles, vegetation, signaling markings, maneuvers, behavior, and traffic situations.

\subsection{nuImages}
Complementing the nuScenes database, the nuImages \cite{nuimgs} dataset provides a set of 93,000 labeled images - 2D masks and bounding boxes - which include rain, snow, and night driving conditions, essential for autonomous vehicle applications.
\par Foreground objects also have attribute annotations - for instance, whether a motorcycle has a rider, the pose of a pedestrian, the activity of a vehicle, flashing hazard lights, and even if an animal is flying.

\subsection{SeeingThroughFog}
Proposed in \cite{b25}, it was specifically created to aggregate scenarios with adverse weather conditions.
\par It provides camera, radar, and other sensing modalities such as LIDAR.
\par It has footage captured in more than 10,000 kilometers of navigation, summing up to 13,500 labeled images - 2D and 3D bounding boxes.

\subsection{BLVD}
Proposed in \cite{b20}, it seeks to foster the development of solutions for a deeper understanding of traffic scenes. To this end, it provides a platform for the tasks of 4D dynamic tracking (3D + temporal), interactive 5D event recognition (4D + interactive), and intent prediction.
\par The dataset provides 120,000 labeled images - 3D bounding boxes -, captured with high and low object density, and under different lighting conditions.
\par Although very powerful in the sense of predicting dynamic and interactive events, the base does not provide data from radar or stereo cameras.

\subsection{Waymo Open Dataset}
Proposed in 2019 \cite{b43}, it contains a total of 200,000 images labeled with 2D bounding boxes, allowing the development of solutions for object detection and tracking.
\par Data acquisition was carried out in different cities, considering different climates, lighting conditions, and navigation contexts - construction sites, dense traffic.
\par Despite being continuously updated and promoting several challenges, it still does not provide radar data.

\subsection{Astyx HiRes2019}
It is a radar-centric automotive dataset designed for 3D object detection and with the objective of fostering the research on radar-based detection and low-level sensor fusion development.
\par Besides radar, it also provides data from camera and LIDAR sensors.

\subsection{H3D}
The Honda 3D Dataset (H3D) \cite{h3d} is a large-scale dataset comprised of 3D LIDAR point clouds and RGB images, all annotated with 3D bounding boxes in order to stimulate the research on 3D object detection and tracking.

\subsection{A2D2}
The Audi Autonomous Driving Dataset (A2D2) \cite{a2d2} was proposed to support startups and academic researchers working on autonomous driving. The dataset features 2D semantic segmentation, 3D point clouds, 3D bounding boxes, and vehicle bus data.
\par Considering all the tasks covered, the dataset has a total of more than 440,000 frames, from which more than 40,000 are labeled with semantics, 12,000 with 3D bounding boxes.

\subsection{A*3D Dataset}
Proposed in 2019, the A*3D dataset seeks to, in the authors' words, push the boundaries of tasks in autonomous driving research to more challenging highly diverse environments \cite{a*3d}. 
\par It has a crucial preoccupation with gathering data from diverse scenes, times (approximately 3 times more night-time images than nuScenes), and weather.

\subsection{EuroCity Persons}
The EuroCity Persons dataset provides a large number of highly diverse, accurate and detailed annotations of pedestrians, cyclists and other riders in urban traffic scenes. The images for this dataset were collected on-board, and cover a great variety of lighting and weather conditions. 
\par In total, there are over 47,000 frames annotated with 2D bounding boxes and orientation.

\subsection{Lyft Level AV Dataset}
Created with the objective of popularizing the use of point clouds with Deep Learning for object detection, the Lyft Level 5 dataset \cite{lyft} provides synchronized LIDAR point clouds and camera images, accounting for more than 350 sequences, each one 60 to 90 minutes-long.
\par Each sequence is labelled with 3D bounding and associated attributes, such as velocity, acceleration, yaw, yaw rate, and a class label.
\par No mention is made to data capture under challenging weather or lighting conditions.

\subsection{Argoverse}
Proposed in 2019 by the Argo AI, the Argoverse dataset \cite{argoverse} was designed in order to foster research on 3D object tracking and motion forecasting. It has a total of 113 sequences of 15 to 30 seconds each, all with 3D bounding box annotations. 
\par The data acquisition was performed under various lighting and weather conditions.

\subsection{PandaSet}
Public large-scale dataset for autonomous driving provided by Hesai \& Scale, the PandaSet \cite{pandaset} enables researchers to study challenging urban driving situations using the full sensor suit of a real self-driving-car.
\par The scenes were selected in order do cover a variety of times of day and lighting conditions in the morning, afternoon, dusk and evening.

\subsection{StreetHazards}
The StreetHazards initiative \cite{streethazards} leverages a simulated driving environment to create a dataset for anomaly segmentation.
\par It contains 5,125 training images, 1,500 test images, and 250 anomaly types.
\par The data was gathered under different (simulated) lighting and weather conditions.

\subsection{Brno Urban Dataset}
Recorded in Brno, Czech Republic, the Brno Urban Dataset comprises sequences accounting for 10h of driving. 
\par It provides data from, among other sensors, four RGB cameras and one infrared camera.

\subsection{Canadian Adverse Driving Conditions Dataset (CADC)}
The Canadian Adverse Driving Conditions (CADC) dataset aims to promote research to improve self-driving in adverse weather conditions. 
\par It is the first public dataset to focus on real-world driving data in snowy weather conditions.

\subsection{Combined Anomalous Object Segmentation (CAOS)}
According to the authors, the Combined Anomalous Object Segmentation (CAOS) Dataset \cite{CAOS} introduces a new benchmark for anomaly segmentation (out-of-distribution detection).
\par It combines two datasets in order to cover both real world, and simulated scenes. The first one is the StreetHazards dataset, created with simulated data from the CARLA simulator \cite{carla}. The second is the BDD-Anomaly dataset, created by sampling the BDD100k dataset, which gathers real-world data, and treating the less frequent classes as anomalies.
\par In total, the CAOS dataset accounts for over 15,000 labeled frames - anomaly masks.

\subsection{Berkeley Deep Drive (BDD100K)}
First announced in 2018, the Berkeley Deep Drive (BDD100K) dataset was proposed to foster research on Heterogeneous Multitask Learning. 
\par It contains over 100K videos, each of which 40 seconds-long, with diverse kinds of annotations including object bounding boxes, drivable areas, lane markings, and full-frame semantic and instance segmentation.
\par Besides very complete in terms of the tasks covered, the BDD100k is also complete in terms of driving conditions represented. Its recordings comprise multiple cities in the United States, under multiple weathers and at different times of the day.

\subsection{CARRADA Dataset}
Proposed in 2021, the CARRADA dataset \cite{b23} provides a set of around 7200 synchronized and labeled - bounding boxes and masks - radar and image readings, encompassing the car, pedestrian, and cyclist categories.

\subsection{RaDICaL}
The Radar, Depth, IMU, RGB Camera for Learning (RaDICaL) \cite{b24} is an open dataset that includes around 220,000 FMCW radar measurements, minimally processed, and aligned with RGB-D images - both labeled.

% \usepackage{multirow}
% \usepackage{graphicx}

\begin{table*}[t]
\caption{Characteristics of the main datasets used for autonomous vehicles' perception.}
\resizebox{\textwidth}{!}{%
\begingroup
\renewcommand{\arraystretch}{1.5} % Default value: 1
\begin{tabular}{cccccccccccc}
\hline
\multirow{2}{*}{Dataset} & \multirow{2}{*}{Year} & \multicolumn{3}{c}{Sensing Modality} & \multicolumn{2}{c}{Adverse Conditions} & \multicolumn{2}{c}{Objective} & \multirow{2}{*}{Size {[}frames{]}*} & \multirow{2}{*}{Link}\\ \cline{3-9} 
                         &                      & 2D       & 3D      & Radar      & Lighting            & Weather           & Detection & Segmentation \\ \hline
KITTI \cite{b17} & 
2012 &
\cmark &
\cmark &
 &   
\cmark &  
 &  
\cmark &  
 \cmark&
\begin{tabular}[c]{@{}c@{}}30,000 (detection)\\
800 (segmentation)\end{tabular}&
https://bit.ly/3CvGRNr
\\ \hline
MS-COCO \cite{ms_coco} &
2015 &
\cmark &
 &
 &   
 &  
 &  
\cmark &  
\cmark &
200,000&
https://bit.ly/3jMNMdG
\\ \hline
KAIST Multispectral Pedestrian \cite{kaist_pedestrian} &
2015 &
\cmark &
 &
 &   
\cmark &  
 &  
\cmark &  
 &
95,328&
https://bit.ly/3CwoFn4
\\ \hline
Udacity \cite{udacity} &
2016 &
\cmark &
 &
 &   
\cmark &  
 &  
\cmark &  
 &
9,423&
https://bit.ly/3jPJd2u
\\ \hline
% Elektra \cite{elektra} &
% 2016 &
% \cmark &
% \cmark &
%  &   
% &  
% &  
% \cmark &  
% \cmark &
% -
% \\ \hline
JAAD \cite{jaad} &
2016 &
\cmark &
 &
 &   
\cmark &  
\cmark &  
\cmark &  
 &
82,032&
https://bit.ly/3mnwOo6
\\ \hline
Tsinghua-Daimler Cyclist Detection Benchmark Dataset \cite{tsinghua} & 
2016 & 
\cmark &
 &
 &   
&  
&  
\cmark &
 &
14,674&
https://bit.ly/3BrKSBg
\\ \hline
Playing for Data: Ground Truth from Computer Games \cite{p4d} & 
2016 & 
\cmark &
 &
 &   
\cmark &  
\cmark &  
 &
\cmark &
25,000&
https://bit.ly/3q4Cwh5
\\ \hline
SYNTHIA \cite{b19} & 
2016 & 
\cmark &
\cmark &
 &   
\cmark &  
\cmark &  
 &
\cmark &
214,000&
https://bit.ly/3jO3fdN
\\ \hline
Cityscapes \cite{b18} & 
2016 &    
\cmark &
 &
 &   
 &  
 &  
\cmark &
\cmark &
25,000&
https://bit.ly/3pV2cN3
\\ \hline
% Oxford RobotCar \cite{oxford_robotcar} & 
% 2016 &
% \cmark&    
% &       
% &            
% \cmark&       
% \cmark&         
% & 
% &
% &
% https://bit.ly/3w0RlSK
% \\ \hline
Multi-spectral Object Detection dataset \cite{MODT} & 
2017 &
\cmark&    
&       
&            
\cmark&       
&         
\cmark& 
&
7,512&
https://bit.ly/3nrMXYM
\\ \hline
Multi-spectral Semantic Segmentation dataset \cite{MSSD} & 
2017 &
\cmark&    
&       
&            
\cmark&       
&         
& 
\cmark&
1,569&
https://bit.ly/3nrMXYM
\\ \hline
Mapillary Vistas \cite{mapillary} & 
2017 &
\cmark&    
&       
&            
\cmark&       
\cmark&         
& 
\cmark&
25,000&
https://bit.ly/3jPdxdk
\\ \hline
KAIST \cite{kaist} & 
2018 &
\cmark&    
\cmark&       
&            
\cmark&       
&         
\cmark& 
&
7,512&
https://bit.ly/3pPw7G9
\\ \hline
ApolloScape \cite{apollo} &
2018 &
\cmark&    
\cmark&       
&            
\cmark&       
\cmark&         
& 
\cmark&
146,997&
https://bit.ly/3nJIhO2
\\ \hline
nuScenes \cite{b21} &
2019 & 
\cmark &
 &
\cmark &   
\cmark &  
\cmark &  
\cmark &
 &
40,000&
https://bit.ly/3nDJgzC
\\ \hline
nuImages \cite{nuimgs} &
2019 & 
\cmark &
 &
 &   
\cmark &  
\cmark &  
\cmark &
\cmark &
93,000&
https://bit.ly/3kYqymB
\\ \hline
SeeingThroughFog \cite{b25} &
2019 &
\cmark &
 &
\cmark &   
\cmark &  
\cmark &  
\cmark &
 &
13,500&
https://bit.ly/3jOT4Wp
\\ \hline
BLVD \cite{b20} & 
2019 &
\cmark &
 &
 &   
\cmark &  
 &  
\cmark &
 &
120,000&
https://bit.ly/3pXO7y2
\\ \hline
Waymo Open Dataset \cite{b43} & 
2019 &
\cmark &
 &
 &   
\cmark &  
\cmark &  
\cmark &
 &
200,000&
https://bit.ly/3w4uPIv
\\ \hline
Astyx HiRes2019 \cite{astyx} & 
2019 &
\cmark &
 &
\cmark &   
&  
&  
\cmark &
 &
500&
https://bit.ly/3CvgQ0C
\\ \hline
H3D \cite{H3D} & 
2019 &
\cmark &
 &
 &   
&  
&  
\cmark &
 &
27,721&
https://bit.ly/3nJXsH6
\\ \hline
A2D2 \cite{A2D2} & 
2019 &
\cmark &
 &
 &   
 &  
\cmark &  
\cmark &
\cmark &
\begin{tabular}[c]{@{}c@{}}41,280 (segmentation)\\
12,499 (detection)\end{tabular}&
https://bit.ly/3pXQO2C
\\ \hline
A*3D Dataset \cite{A*3D} & 
2019 &
\cmark &
 &
 &   
\cmark &  
\cmark &  
\cmark &
 &
39,000&
https://bit.ly/3jNmnZc
\\ \hline
EuroCity Persons \cite{eurocity} & 
2019 &
\cmark &
 &
 &   
\cmark &  
\cmark &  
\cmark &
 &
47,300&
https://bit.ly/3pQKkmq
\\ \hline
Lyft Level 5 AV Dataset 2019 \cite{lyft} & 
2019 &
\cmark &
 &
 &   
&  
&  
\cmark &
 &
\begin{tabular}[c]{@{}c@{}}More than 350 recordings\\ (60 to 90 seconds long)\\
525,000 frames (70 seconds long sequences, sampled at 20 Hz)\end{tabular}&
https://bit.ly/3bm7Cbd
\\ \hline
Argoverse \cite{argoverse} & 
2019 &
\cmark &
\cmark &
 &   
\cmark &  
\cmark &  
\cmark &
 &
\begin{tabular}[c]{@{}c@{}}113 recordings\\ (15 to 30 seconds long) \\
50,850 frames (22.5 seconds long sequences, sampled at 20 Hz)\end{tabular}&
https://bit.ly/3GCFDSX
\\ \hline
PandaSet \cite{pandaset} & 
2019 &
\cmark &
 &
 &   
\cmark &  
 &  
\cmark &
 &
48,000&
https://bit.ly/3GxjjKt
\\ \hline
StreetHazards \cite{streethazards} & 
2019 &
\cmark &
 &
 &   
\cmark &  
\cmark &  
 &
\cmark &
7,656&
https://bit.ly/3bm8wED
\\ \hline
Brno Urban Dataset \cite{brno} & 
2019 &
\cmark &
 &
 &   
\cmark &  
\cmark &  
\cmark &
 &
\begin{tabular}[c]{@{}c@{}}67 recordings\\ (summing up to 10h)\\
720,000 frames (frame rate of 20 Hz)\end{tabular}&
https://bit.ly/3GBoycg
\\ \hline
% Oxford Radar RobotCar \cite{b22} & 
% 2020 &
% \cmark&    
% &       
% \cmark&            
% \cmark&       
% \cmark&         
% & 
% &
% &
% https://bit.ly/3EuyVg1
%\\ \hline
Canadian Adverse Driving Conditions Dataset \cite{canadian} & 
2020 &
\cmark&    
&       
&            
&       
\cmark&         
\cmark& 
&
7,000&
https://bit.ly/2XWcoJm
\\ \hline
Combined Anomalous Object Segmentation (CAOS) \cite{CAOS} & 
2020 &
\cmark&    
&       
&            
\cmark&       
\cmark&         
& 
\cmark&
\begin{tabular}[c]{@{}c@{}}7,656 (simulated)\\ 8,000 (real)\end{tabular}&
https://bit.ly/3bm8wED
\\ \hline
Berkeley Deep Drive (BDD100K) \cite{bdd100k} & 
2020 &
\cmark&    
&       
&            
\cmark&       
\cmark&         
\cmark& 
\cmark&
\begin{tabular}[c]{@{}c@{}}100,000 (detection)\\
20,000 (segmentation)\end{tabular}&
https://bit.ly/3jREwoL
\\ \hline
CARRADA \cite{b23} & 
2021 &
\cmark&    
&       
\cmark&            
\cmark&       
\cmark&         
\cmark&     
\cmark&
7,200&
https://bit.ly/2ZyULjg
\\ \hline
RaDICaL \cite{b24} & 
2021 &
\cmark&    
\cmark&       
\cmark&            
&       
&         
\cmark **&
\cmark **&
220,000(check)&
https://bit.ly/3BwT42Z
\\ \hline
\end{tabular}%
\endgroup
}
\label{table:1}
\caption*{* When not explicitly mentioned otherwise. ** Although the authors mention the possibility of using the dataset for this purpose, no labeled data is provided.}
\end{table*}

\begin{figure}
\centering
\includegraphics[width=\columnwidth]{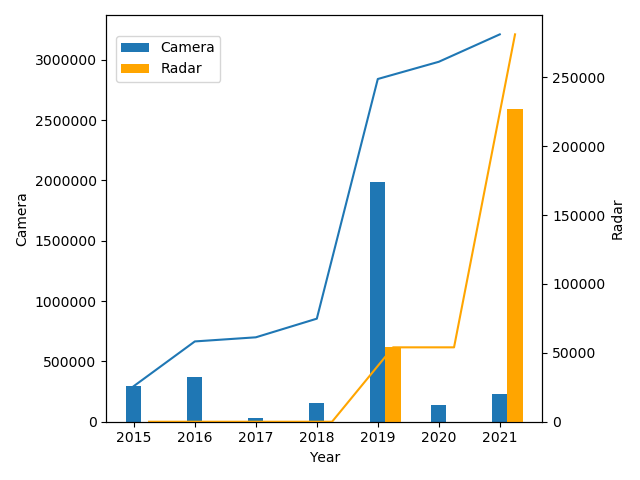}
\caption{Availability of radar and camera labeled data aimed at detection and segmentation in the context of autonomous vehicles and ADAS, from 2015 to 2021.}
\label{fig:evol_mod}
\end{figure}

\section{Metrics}
The set of metrics used for evaluating the performance of detection and segmentation methods is extremely vast. The main ones are listed below.
\par Often, such measures are defined along with the proposition of the databases. Hence, the they are calculated can vary according to the benchmark being considered.
\par Table \ref{table:metrics} summarizes the metrics used by each of the databases considered in this review (section \ref{datasets}).

\subsection{Detection}
Detection tasks can be divided into 2D detection and 3D detection, as defined in section \ref{detection}.
\par In 2D detection, metrics are often used to assess classification accuracy, detection center point alignment, and the overlap of generated and ground truth bounding boxes.
\par In the first case, False Positives per Image ($FP_{i}$), Accuracy (AC [\%]) and Average Precision (AP) can be used. The miss rate (MR), as well as some derivations as the Log-Average Miss Rate(LAMR) \cite{eurocity, kaist_pedestrian}, are also commonly used metrics.
\par To analyze the alignment between the ground-truth and the detected center points, the Average Translation Error (ATE) can be used. It can be calculated in pixels or meters, depending on the coordinate system used - image or world.
\par To evaluate the overlap between the generated bounding boxes, the Intersection over Union (IoU) can be used.
\par When considering 3D object detection, several attributes can be added to the ones observed in 2D detection. For example, to a 3D bounding box can be associated its orientation, speed, and attributes linked to the action being performed by the entity under analysis.
\par Thus, in the case of 3D detection, in the same way as done for 2D detection, Average Precision can be used to evaluate the matching between the generated boxes. The measure termed Average Precision Weighted by Heading (APH) \cite{b43} can also be used in this context.
\par The center alignment of the generated boxes, their orientation, scale (volume alignment after translation and orientation alignment), speed and other action attributes can be evaluated by Average Translation Error (ATE), Average Scale Error (ASE), Average Orientation Error (AOE), Average Velocity Error (AVE) and Average Attribute Error (AAE), respectively. Such metrics are defined in \cite{b21}, where they are named True Positive Metrics.
\par An analogous way to assess orientation is through Average Orientation Similarity (AOS) \cite{b17}.

\subsection{Segmentation}
When evaluating semantic segmentation, the Intersection over Union (IoU), as well as derived metrics, such as the IoU value per class (\(IoU_{class}\)), per category (\(IoU_{category}\), and at the instance level (iIoU) \cite{b17} are generally used. The analysis can also be done in terms of Pixel Precision (PP) and Pixel Recall (PR) \cite{b23}. It is also possible to evaluate the segmentation performance through the Class Average Accuracy (classAvg) \cite{MSSD}. Finally, some works apply the Area Under the ROC Curve (AUC), the Area Under the precision-recall curve (AUPR), and the False Positive Rate at different recall levels ($FP^{\%}$) \cite{streethazards, CAOS}.
\par To evaluate the performance of instance segmentation methods, the Average Precision (AP) can be applied. Additionally, many variations of the AP are proposed in the literature. It can be calculated for different values of overlap \cite{b17} - the average precision for detections with 50\% of overlap is defined as \(AP_{50\%}\) -, different IoU values \cite{ms_coco} - \(AP^{IoU=0.50}\) for a IoU value of 0.5  -, different distance measures \cite{b18} - \(AP_{50m}\) for objects at 50 meters - and different scales \cite{ms_coco} - and for different instance sizes - small (\(AP^{small} \)), medium (\(AP^{medium}\)) and large (\(AP^{large}\)) objects.
\par It is also possible to evaluate the Average Recall (AR), given different number of detections per image. For example, \cite{ms_coco} evaluates the Average Recall for thresholds of up to 1, 10 and 100 detections per image - defined as \(AR^{max=1}\), \(AR^{max=10}\) and \(AR^{max=100}\), respectively.
\par Finally, to evaluate the performance of panoptic segmentation methods, the panoptic quality (PQ) is the standard metric \cite{panoptic_seg}.

\subsection{General Metrics}
In addition to the previous metrics, more general performance measures can be considered. In the of autonomous vehicles, the inference rate, memory footprint, computational complexity, and model size are of great relevance.

\begin{table*}
\caption{Metrics for performance evaluation of perception methods, for each dataset.}
\begin{adjustbox}{max width=\textwidth}
\begingroup
\renewcommand{\arraystretch}{1.5} % Default value: 1
\begin{tabular}{ccccc}
\hline
\multirow{2}{*}{Dataset} &
  \multicolumn{2}{c}{Metrics}  \\ \cline{2-3}
 &
  Detection &
  Segmentation &
   \\ \hline
KITTI \cite{b17} &
  \begin{tabular}[c]{@{}c@{}}AP, AOS\end{tabular} &
  \begin{tabular}[c]{@{}c@{}}AP, $AP^{50\%}$,\\ $IoU_{class}$, $iIoU_{class}$,\\ $IoU_{category}$, $iIoU_{category}$\end{tabular}  \\ \hline
MS-COCO \cite{ms_coco} &
  \begin{tabular}[c]{@{}c@{}}AP, $AP^{IoU=.50}$,\\ $AP^{IoU=.75}$, $AP^{small}$,\\ $AP^{medium}$, $AP^{large}$,\\ $AR^{max=1}$, $AR^{max=10}$,\\ $AR^{max=100}$, $AR^{small}$,\\ $AR^{medium}$, $AR^{large}$\end{tabular} &
  \begin{tabular}[c]{@{}c@{}}mIoU, fIoU, mAcc,  pAcc, PQ\end{tabular} \\ \hline
KAIST multispectral pedestrian \cite{kaist_pedestrian} &
  Recall, $LAMR_{all}$, $LAMR_{day}$, $LAMR_{night}$&
  
  \\ \hline
Udacity \cite{udacity} &
  &
  \\ \hline
% Elektra \cite{elektra} &
%   &
%   &
%   \\ \hline
JAAD \cite{jaad} &
  AP, mAP &
  \\ \hline
Tsinghua-Daimler Cyclist Detection Benchmark Dataset   \cite{tsinghua} &
  AP &
  \\ \hline
Playing for Data: Ground Truth from Computer Games \cite{p4d} &
  &
  mIoU \\ \hline
SYNTHIA \cite{b19} &
  &
  mIoU  \\ \hline
Cityscapes \cite{b18} &
  AP &
  \begin{tabular}[c]{@{}c@{}}AP, $AP^{50\%}$, $AP^{50m}$,\\ $AP^{100m}$, IoU, iIoU, PQ\end{tabular}  \\ \hline
% Oxford RobotCar \cite{oxford_robotcar} &
%   &
%   \\ \hline
Multi-spectral Object Detection dataset \cite{MODT} &
  mAP &
  \\ \hline
Multi-spectral Semantic Segmentation dataset \cite{MSSD} &
  &
  mIoU, \textit{classAvg} \\ \hline
Mapillary Vistas \cite{mapillary} &
  &
  mIoU, $AP_{class}$, mAP\\ \hline
KAIST \cite{kaist} &
  \textit{missRate} &
  \\ \hline
ApolloScape \cite{apollo} &
  &
  \begin{tabular}[c]{@{}c@{}}
  mIoU, $Acc_{pix}$, $mAcc_{class}$, \\ 
  mAP, $mAP_{class}$\end{tabular} \\ \hline
nuScenes \cite{b21} &
  \begin{tabular}[c]{@{}c@{}}mAP, mATE, mASE,\\ mAOE, mVE, mAAE\end{tabular} &
   \\ \hline
nuImages \cite{nuimgs} &
  &
  \\ \hline
SeeingThroughFog \cite{b25} &
  AP &
 \\ \hline
BLVD \cite{b20} &
  &
  \\ \hline
Waymo Open Dataset \cite{b43} &
  \begin{tabular}[c]{@{}c@{}}AP, APH\end{tabular} &
   \\ \hline
Astyx HiRes2019 \cite{astyx} &
  AP &
  \\ \hline
H3D \cite{H3D} &
  mAP &
  \\ \hline
A2D2 \cite{A2D2} &
  &
  mIoU\\ \hline
A*3D Dataset \cite{A*3D} &
  mAP &
\\ \hline
EuroCity Persons \cite{eurocity} &
  \begin{tabular}[c]{@{}c@{}}LAMR, $FP_{i}$\end{tabular} &
\\ \hline
Lyft Level 5 AV Dataset 2019 \cite{lyft} &
  mAP &
 \\ \hline
Argoverse \cite{argoverse} &
  &
\\ \hline
PandaSet \cite{pandaset} &
  &
\\ \hline
StreetHazards \cite{streethazards} &
  &
  \begin{tabular}[c]{@{}c@{}}AUC, $FP^{95\%}$, AUPR\end{tabular}\\ \hline
Brno Urban Dataset \cite{brno} &
  &
\\ \hline
% Oxford Radar RobotCar \cite{b22} &
%   &
% \\ \hline
Canadian Adverse Driving Conditions Dataset \cite{canadian} &
  &
\\ \hline
Combined Anomalous Object Segmentation (CAOS) \cite{CAOS} &
  &
  \begin{tabular}[c]{@{}c@{}}AUC, $FP^{95\%}$, AUPR\end{tabular}\\ \hline
Berkeley Deep Drive (BDD100K) \cite{bdd100k} &
  AP &
  mIoU \\ \hline
CARRADA \cite{b23} &
  &
  \begin{tabular}[c]{@{}c@{}}mIoU, mPP, mPR\end{tabular} \\ \hline
RaDICaL \cite{b24} &
  AC {[}\%{]} &
\\ \hline
\end{tabular}
\endgroup
\end{adjustbox}
\label{table:metrics}
\end{table*}

\section{Challenges and Open Questions}
In the following subsections, we discuss the main challenges and open questions observed during the present review.

\subsection{ADAS and Autonomous Vehicles}

\par Advanced Driver Assistance Systems, being a more mature technology and a reality nowadays - most of manufacturers already deliver some degree of assistive technologies in their vehicles -, witness an also more mature regulation \cite{euroncap, latinncap}.
\par In the case of Autonomous Vehicles, however, the legislation is still under development and testing parallel to the development of the fully automated cars themselves. 
\par In developed countries, such as the United States, the tests on Autonomous Vehicles have been made at an accelerated rate. The National Highway Traffic Safety Administration (NHTSA) provides a tool to keep track of these tests, which is part of their Automated Vehicle Transparency and Engagement for Safe Testing Initiative \cite{nhtsa_av_test}. The Society of Automotive Engineers is also another organization working towards clarifying the field and making it possible to propose a more precise regulation \cite{sae}.
\par In low-incoming countries, however, this process is considerably slower. The different transit rules, road conditions, and even the lack of investment, can impact in the development of legislation regarding Autonomous Vehicles - or even the adoption of a legislation brought from another country, more advanced in this subject.

\subsection{Sensing modalities}
Deep Learning-based radar perception remains under-explored, offering much room for advances. 
\par One practical evidence of that is the low number of radar and camera-radar fusion-based approaches submitted to the challenges and benchmarks related to the datasets covered by our work. LIDAR-based methods are predominant, despite its considerably low accessibility in terms of price.
\par One possible reason for that is the lack of benchmarks specifically aimed at radar-based detection. ROD21 \cite{ROD2021} was a pioneer in this sense, and the great number of submissions received - in the order of hundreds of works - show that the community is willing to develop radar-based perception approaches.
\par Initiatives like that also contribute to tackling the lack of large-scale open-source datasets, as many of the radar datasets available to date are only a fraction of proprietary ones. In addition, it helps to increase the availability of labeled radar data, what, as shown in figure \ref{fig:evol_mod}, is a tendency since 2019.
\par A last open question is the use of radar in segmentation and detection. Many of the works in this direction are based on designs proposed for either for LIDAR point cloud processing, or for RGB image processing. Hence, possible advantages of radar characteristics may remain unexplored by the current Deep Learning architectures.
\par When considering camera-based perception, although being the most mature sensing modality, there are also many open questions. First of all, despite of providing high resolution and rich color, shape and texture information, plain RGB cameras lack of a crucial information in environment perception, which is depth. Stereo Vision fills this need, but just as monocular vision, suffers with degradation caused by challenging operation conditions.
\par A current tendency observed is this direction is the use of multi-spectral data, captured with infra-red cameras, since they provide a more robust perception in challenging lighting conditions. However, operation in adverse weather conditions remains an open challenge that could possibly be solved through the used of data fusion approaches.
\par Finally, when it comes to data fusion approaches, the main challenge relates to finding the best answers to the questions of ''what to fuse'', ''how to fuse'' and ''when to fuse''. An architecture-agnostic method for data fusion still remains an open question.

\subsection{Perception Tasks}
Data labeling, besides being an error-prone process when made by human annotators, is a very costly activity and, sometimes, even impractical. In video sequences, for example, it is common that not every frame is labeled \cite{b17, b18}. Therefore, one challenge in supervised Deep Learning-based object detection and, principally, in supervised Deep Learning-based image segmentation is developing methods for learning even with sparse annotations. One approach in this direction is to use label propagation through optical flow \cite{opt_flow_5}.
\par Another open question refers to the application radar data in instance and panoptic image segmentation.

\subsection{Datasets}
The review presented in this work elucidates the growing concern with the development of methods robust to challenging conditions of perception, in the context of autonomous vehicles.
\par One of the strongest evidences of this trend is the construction of datasets covering the widest possible variety of factors that can influence in vehicle perception and navigation. Some examples are weather - snow, rain, fog -, lighting - sunrise, sunset, night - and traffic conditions - road works, dense traffic.
\par Another evidence pointing to this direction is the growing presence of data from different modalities, other than just images, in the recently proposed databases - in fact, if we take into account that autonomous vehicles carry a multitude of sensors, such as camera, radar, LIDAR, and GPS, this is a natural evolution. One sensing modality that has gained relevance is based radar perception, as figure \ref{fig:evol_mod} illustrates. From 2019 onwards, besides many datasets specifically designed for radar-based perception have been proposed, camera datasets have also reserved an important fraction of its size for radar readings.
\par Another issue in terms of datasets refers to data labeling. Principally in datasets designed to foster the development in image segmentation, data annotation is a very time-consuming process, what can explain the considerably less presence of datasets aimed at segmentation when compared to the ones designed for object detection. This phenomenon was termed as the curse of dataset annotation in \cite{curse_of_annotation}. Some works try to address this problem by proposing alternative labeling methods \cite{p4d, curse_of_annotation}.
\par Another challenge refers to the coverage of different driving contexts. As shown in \cite{survey1}, most of the data available were gathered in developed countries, such as the United States, and countries from Europe, and Asia, where the development of autonomous vehicles is advanced compared to other regions in the globe. Therefore, low-income countries, where the traffic environment greatly varies from well structured to not structured at all - rural roads with no clear signalization -, lack representation. Relevant efforts in this direction have been made in \cite{mapillary}.
\par
\begin{figure*}
  \includegraphics[width=\textwidth]{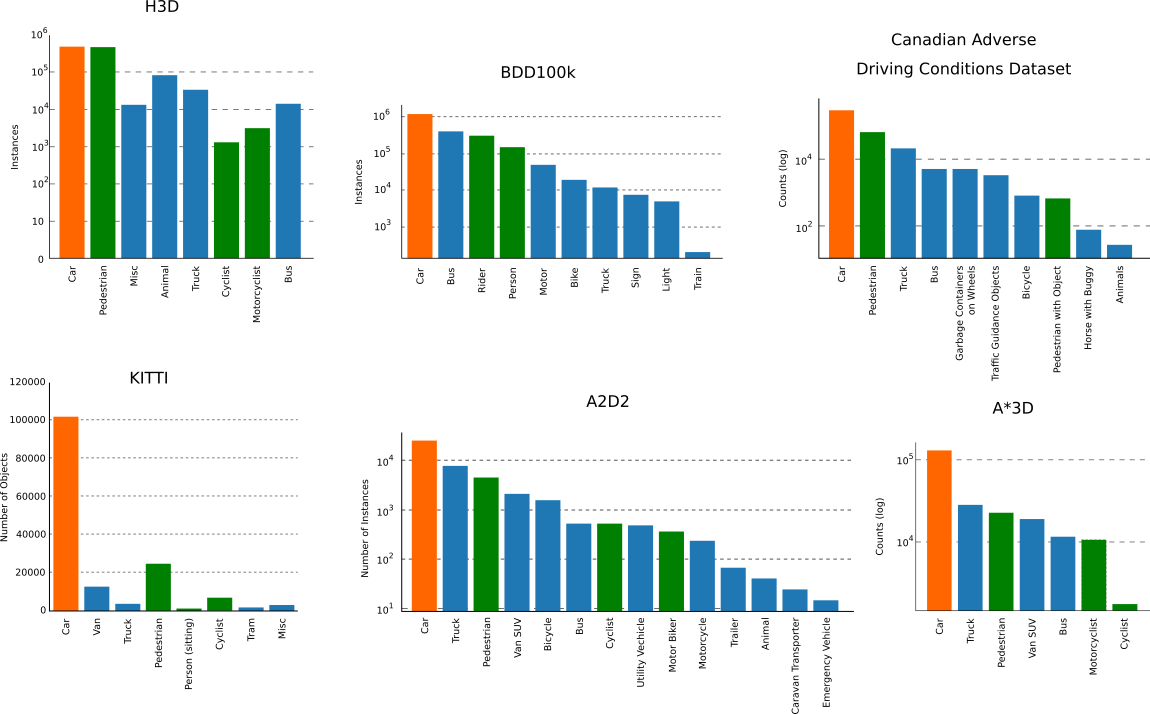}
  \caption{The problem of data unbalance encountered in many the large-scale datasets for vehicle perception.}
  \label{fig:data_unbalance}
\end{figure*}

 The last challenge refers to data imbalance in many datasets. As shown in Figure \ref{fig:data_unbalance}, the class car is predominant to other classes of utmost importance in vehicle perception, such as the ones related to vulnerable road users - represented in green in figure \ref{fig:data_unbalance}.
Some datasets already focus their attention in solving this issue, delivering a less skewed class distribution \cite{h3d}. Other works are specifically aimed a certain classes, as is the case of datasets designed for the detection of vulnerable road users \cite{kaist_pedestrian, jaad, tsinghua, eurocity}.
Another possible solution is to use loss functions robust to class imbalance.

\subsection{Metrics}
The last challenge identified is the lack of standardization in terms of the metrics used to evaluate the perception for autonomous vehicles and ADAS. Such metrics are generally defined to apply regardless of the application. Additionally, they may vary according to the dataset - different benchmarks may use different definitions -, making it difficult to perform comparative analysis.
\par One possible solution is to develop and use metrics aligned with the main regulations \cite{euroncap, latinncap} since this can make it easier to assess the methods in the specific context of ADAS and Autonomous Vehicles. For instance, the mAP could be evaluated for different distances from the ego vehicle \cite{b18}, depending on the context of navigation. In urban scenarios, high precision at distances far from the vehicle is less critical, since the speed is considerably lower than on the road.

\section{Conclusion}
The present study evaluates the current research scenario in detection and segmentation for Autonomous Vehicles and Advanced Driver Assistance Systems (ADAS).
\par Initially, we presented some fundamental concepts related to ADAS, Autonomous Vehicles, and the path towards fully automated driving.
\par Then, we introduced the main concepts and the current context on camera and radar-based perception. We highlighted the importance of considering sensor fusion techniques for robust perception under challenging weather and lighting conditions.
\par Next, we presented the main concepts related to Deep Learning-based detection and segmentation.
\par This was followed by an analysis of the current datasets used to foster the research on autonomous vehicles and ADAS. We covered different characteristics, such as the data modalities provided, the coverage of adverse navigation conditions, their size, and purpose.
\par Then, we discussed the most common metrics used to assess the performance of detection and segmentation methods.
\par Finally, we presented some of the main challenges and open questions in the field.
\par The results of this review point to a growing concern with the development of vehicle perception methods robust to challenging operating conditions. This is represented by the increasing availability and use of radar and data fusion-based approaches. However, those approaches remain under-explored, with much room for improvement in terms of accuracy and efficiency.
\par Additionally, the adoption of standard metrics for evaluating the performance of perception methods, aligned with regulatory attempts in the context of ADAS and autonomous vehicles, may help accelerate the development of the field.
\par Finally, we highlight the importance of providing balanced and diverse data for the development of robust assistance and autonomous systems. The training of Deep Learning models to recognize under-represented classes, such as cyclists, and to deal with challenging conditions, such as diverse traffic infrastructure, weather, and lighting conditions is of utmost importance towards road safety in the future of transportation.

\end{document}